\documentclass[aps,prx,reprint,amsmath,amssymb,superscriptaddress,longbibliography]{revtex4-1}%,linenumbers <-- add "linenumbers" as option above if desired
\usepackage{graphicx}% Include figure files
\usepackage{subcaption}% Subfigures
\usepackage{dcolumn}% Align table columns on decimal point
\usepackage{bm}% bold math
\usepackage{amsmath}
\usepackage{amssymb}
\usepackage{xcolor}
\usepackage{xspace}
\usepackage{xfrac}
\usepackage{color,soul}

\begin{document} 

\title{Hebbian learning the local structure of language}
\author{P. Myles Eugenio}
\email{pamyeuge@iu.edu}
\affiliation{Department of Physics, Indiana University, Bloomington, Indiana 47405, USA}
\affiliation{Department of Physics, University of Connecticut, Storrs, Connecticut 06269, USA}
\affiliation{Department of Physics, Harvard University, Cambridge, Massachusetts 02138, USA}

\begin{abstract}
Learning in the brain is local and unsupervised (Hebbian). We derive the foundations of an effective human language model inspired by these microscopic constraints. It has two parts: (1) a hierarchy of neurons which learns to tokenize words from text (whichiswhatyoudowhenyoureadthis); and (2) additional neurons which bind the learned symanticless patterns of the tokenizer into a symanticful token (an embedding). The model permits continuous parallel learning without forgetting; and is a powerful tokenizer which performs renormalization group. This allows it to exploit redundancy, such that it generates tokens which are always decomposable into a basis set (e.g an alphabet), and can mix features learned from multiple languages. We find that the structure of this model allows it to learn a natural language morphology {\it without} data. The language data generated by this model predicts the correct distribution of word-forming patterns observed in real languages, and further demonstrates why microscopically human speech is broken up into words. This model provides the basis for understanding the microscopic origins of language and human creativity.

\end{abstract}
\maketitle

\section{Introduction}

In the late 1970's, deaf Nicaraguan school children invented an indigenous sign language unrelated to any already existing language (spoken or signed) \cite{Kegl&Iwata1989_LSN,Senghas1995_SignLanguage,Senghas&Coppola2001_SignLanguage}. These children had no previous exposure to a developed language. The earliest stages of this language developed spontaneously out of their collective interactions; later becoming increasingly nuanced and systematized as younger incoming generations acquired it from older students \cite{Senghas&Coppola2001_SignLanguage}. 

The formation of Nicaraguan Sign Language (NSL) is perhaps the sharpest example of human learning absent data. Its circumstances are unique to a disability, making its emergence unbiased by existing data. Similar learning of novel language has occurred throughout history, such as in the formation of creoles \cite{reviewCreoles2018,Mcwhorter&Good2012_Saramaccan}. This process of rapid language generation \& change is typically driven by the youngest learners -- a surprising contrast with data-hungry large language models (LLMs).

For contrast, LLMs struggle to learn existing creoles due to the limitations on available data \cite{Lent2021_LanguageModels4Creoles,Lent2024_CreoleBenchmarks,Feldman2020_SupervisedMemorization}. Though modern generative AI have improved impressively in recent years \cite{Wei2023_CoT,Long2023_ToT,LLama3HerdofModels_2024,DeepSeek2025_ReinforcementLearning}, the predicted rate of improvement is power-law in the data and compute \cite{Kaplan2020_ScalingLawsNeuralLanguageModels,Vyas2023_FeatureLearningNetworksConsistentWidths,Hestness2017_DeepLearningScalingPredictable,Bahri2024_ExplainingNeuralScalingLaws,Aghajanyan2023_ScalingLawsMixedModalLLMs}. Training such models is only possible because of the mass recording and centralization of human data, which are curiously not the conditions which led to that data.

This opens a broader question into the existence and origin of language and its data. NSL did not exist in the '60s but existed by the '70s. In other words: How do we go from a universe without language to a universe with language? Dense models which require data in order to produce language cannot account for its existence. From this light, it becomes apparent that NSL and the creoles are the tip of the iceberg: We speak all these languages. Where did all the data come from?

In this work, we provide the essential framework for answering this question. To do this, we reexamine the microscopic conditions of human speech. Here {\it microscopic} meaning in the most detailed sense at the 2-neuron level. At this level, the only justified mechanism for learning is Hebbian learning, which describes the tendency for neurons to correlate their firing. In the language of the Hopfield networks \cite{Hopfield1982,HopfieldNeurobiology_Krotov&Hopfield2021}, these correlations between the firing state vectors $u_j$ (with firing rate $v_j\equiv v_j(u_j)$) of different neurons are encoded in a matrix $g_{jk}$, which evolves as 
\begin{eqnarray}\label{Eqn:Hebb}
\tau_g \dot{g}_{jk} + g_{jk} = v_j(u_j)v_k(u_k)
\end{eqnarray}
over timescale $\tau_g$. The right-hand side of Eqn \ref{Eqn:Hebb} can be understood as arising from the gradient descent of a network energy \cite{HopfieldNeurobiology_Krotov&Hopfield2021}, $H=\sum_{jk}v_jg_{jk}v_k$, which encodes the local (2-point) interaction between neurons. Learning in this paradigm is {\it local} and {\it unsupervised}.

Unlike the microscopic scale, the correlations between the tokens of speech are longer than 2. For example, the 3-point correlation {\bf str} in English {\bf strength}. (The tokens here being the alphabetic letters.) Even longer correlations make up the morphology of {\bf strong}, {\bf strongest}, \& {\bf strength}. More generally, speech is composed of correlated strings of arbitrary length ($N$-point). A coherent sentence requires that its letters be correlated with the other letters of their word, as well as with the letters which compose the words elsewhere in the sentence. 

%In the simplest sense, we may describe a language model as a collection of correlated strings, each with a probability of being inferred from a context; where that probability is ordered by the string's reasonableness given that context. 
Long correlations are a defining property of a successful language model, without which it would exhibit a lack of coherence and an inability to stay on topic. This makes understanding why such correlations arise, in spite of the microscopic constraints, a central puzzle in understanding human speech. Large language models are aided in this task by having attention mechanisms with non-linear activation functions, like softmax, which generate non-linear superpositions during inference \cite{Vaswani2018_AttentionAllYouNeed}. The transformer architecture has already been reformulated as a modern Hopfield network \cite{HopfieldAllNeed_Ramsauer2021,EnergyTransformer_Hoover&Krotov2023}, and it has been argued that such models are {\it biologically plausible} because they can be derived from local theories under some assumptions \cite{HopfieldNeurobiology_Krotov&Hopfield2021,Krotov2023_AstrocyteTransformer,Rahimi&Benjamin2007_RandomFeatures,Salinas&Abbott1994_CosineTuningCurve}. This includes allowing for exponential memory capacity for dense autoassociative memories \cite{HopfieldAllNeed_Ramsauer2021,HopfieldNeurobiology_Krotov&Hopfield2021}. However, while these black boxes can successfully learn long correlations, they make no predictions, and do not teach us anything about the microscopic origins of language and its structure. If human language memory was exponential, then a language with only $10$ syllables could communicate $10^5$ messages in a string of length $5$. Instead human speech is broken up into discrete locally-correlated chunks (words), whose entropy for a length $N$ string is less than exponential \cite{Herdan1958_logNormal_intraword,Williams1940_logNormal_interwords,Eckhard&Werner&Markus2001_logNormal_review,MarkNewman2006_powerlaws,Eugenio2023_PhonotacticMemory}.

%If we have exponential memory capacity, then we could communicate with exponentially dense codes, such that a message of size $N$ could be one of $d^N$ valid messages. Instead human speech is broken up into discrete locally-correlated chunks (words), whose entropy for a length $N$ string is less than exponential \cite{Eugenio2023_PhonotacticMemory}.

We then ask if it is possible to learn correlated strings without violating the unsupervised and local constraints of the biology. We find that it is possible with the aid of a hierarchy of local interactions, each of the form Eqn \ref{Eqn:Hebb}. The resulting model learns words by strengthening correlations between neighboring tokens in the text, starting first with letter-letter correlations (bigrams). These learned bigrams are then used to define compound tokens, which are used to train correlations ($n$-grams) at the next level of the hierarchy. This process is repeated, generating a series of projector maps used for tokenization. Such a model is an $n$-gram model \cite{Che&Hall&Jakobson1953_PhonemeGroups} which learns by playing a game akin to byte pair encoding (BPE) \cite{Gage1994_BPE}; except with the additional constraint that all learned $n$-grams be composed only of learned $n-1$ grams, which are the nodes of a directed acyclic graph. The unsupervised and hierarchical graph-forming nature of our model shares similarities with ADIOS \cite{Solan&Horn&Ruppin&Edelman2005_ADIOS}.

We find that when trained against uniformly random strings, or if the hierarchies are grown randomly, the model learns the symanticless patterns of a novel random language morphology. The vocabulary of this random language can be extracted through replay, where a random alphabetic character is provided as context for inference. This random vocabulary is (1) tokenizable, ``whichistosayyoucanreaditlikethis"; (2) has a distribution of unique word-forming $n$-grams which fits well (for a broad choice of model parameters) a log-normal distribution, as previously estimated for real languages \cite{Herdan1958_logNormal_intraword,Williams1940_logNormal_interwords,Eckhard&Werner&Markus2001_logNormal_review,MarkNewman2006_powerlaws}; and (3) exhibits a series of persistent Zipf-like power-laws in the rank-ordered frequency distributions, which is indicative of morphology.

However, we find that our hierarchical model suffers from a combination of forgetting and poor scaling, which prevents it from successfully tokenizing strings of indefinite length. Conveniently, both these problems are resolved if additional neurons, not part of the original hierarchy, are made to fire during the replay of these hierarchies. This replay relearning is completely random \& unsupervised. It has three key effects: (1) an embedding is learned, which ties together all the features ($n$-grams) of a replayed word, as well as the projection maps needed to tokenize those features from basis tokens; (2) these embeddings live in the synaptic connections to the added neurons, so that the information stored in the hierarchical neurons can be forgotten; and (3) makes possible a compression. The resulting compressed set of embeddings are independent from one another, such that both learning and inference can be completely parallelized. Learning can occur continuously without forgetting so long as new neurons are added to the system. The interaction between the tokenizing $n$-gram model and the added (embedding) neurons gives rise to a key-value memory \cite{Gersham&Fiete&Irie2025_KeyValueMemory}, which allows for the fast recognition of words in a string.

This paper is organized as follows: The model is introduced in Sec \ref{Sec:model}. We explore how the model tokenizes the text, by using a minimal example, in Sec \ref{Sec:retokenization}. We discuss how the model scales during training in Sec \ref{Sec:explodingdimensions}. We introduce replay relearning and compression in Sec \ref{Sec:replay}. In Sec \ref{Sec:ralang}, we use the model to generate a vocabulary for a random language by randomly growing the hierarchies. In Sec \ref{Sec:neuralmorphology} \& \ref{Sec:randomprojections} we predict the existence of a tokenizable neural code, and discuss its experimental signature.

%\begin{figure}
%    \centering
%    \includegraphics[width=1.\linewidth]{featuremaps.png}
%    \caption{A hierarchy is defined by a sequence of Hamiltonians related by projectors. The features at each level of the hierarchy are learned tokens ($\tilde{v}$) representing $n$-grams. Below it we show the tokens generated during each step of replay, and the contribution to the Hamiltonian which binds those features to an embedding vector ($a$).}
%    \label{Fig:featuremaps}
%\end{figure}

\begin{figure}
\centering
\begin{subfigure}[t]{0.5\textwidth}
    \centering
    \includegraphics[width=.9\linewidth]{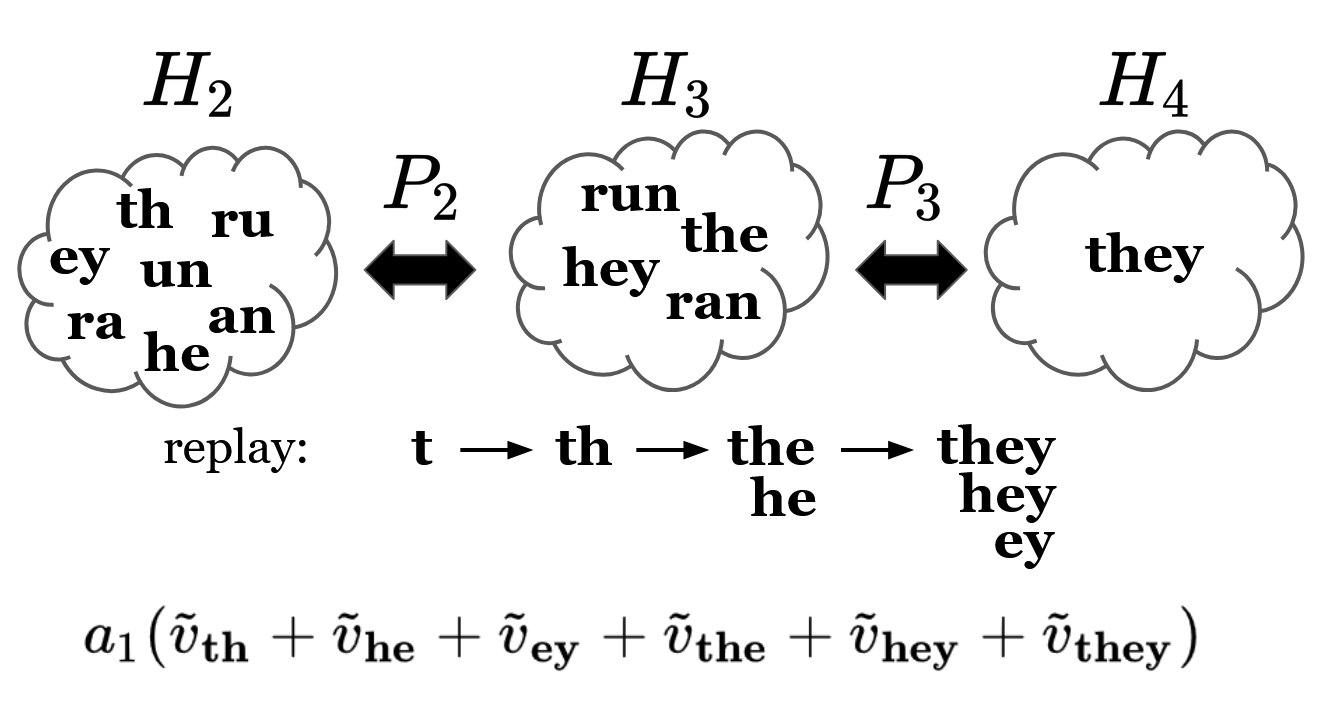}
    \caption{A hierarchy is defined by a sequence of Hamiltonians related by projectors. The features at each level of the hierarchy are learned tokens ($\tilde{v}$) representing $n$-grams. Below it we show the tokens generated during each step of replay, and the contribution to the Hamiltonian which binds those features to an embedding vector ($a$).}
    \label{Fig:featuremaps}
\end{subfigure}\\
\begin{subfigure}[t]{0.5\textwidth}
    \centering
    \includegraphics[width=.9\linewidth]{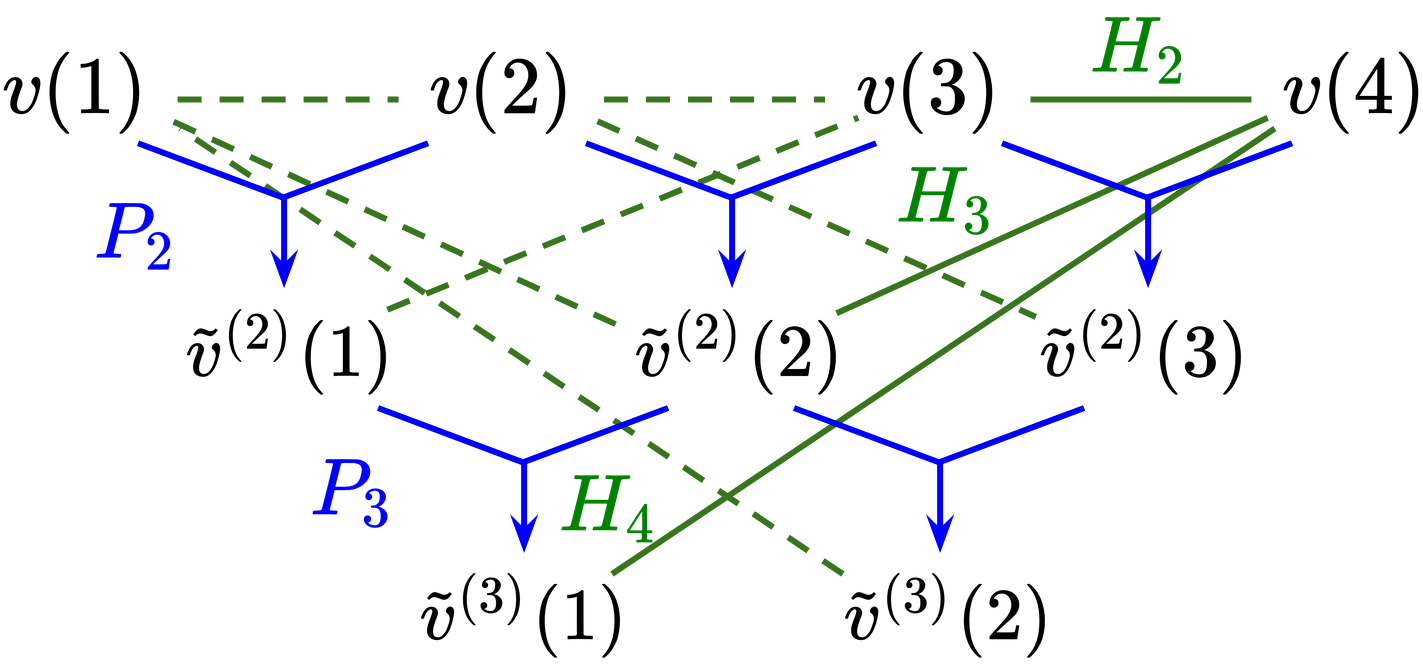}
    \caption{Segment of the uniform hierarchical chain. Solid lines labeled by $H_n$ show how projections to learned features allow $v(x<4)$ to correlate with $v(4)$.}
    \label{Fig:hierarchicalchain}
\end{subfigure}
\caption{}
\end{figure}

%\begin{eqnarray}\label{Eqn:IsingN_tilde}
%H(x) \notag 
%&=& \sum\limits_k\Psi_{k}(x)f_k(v_k(x)) + \sum\limits_{j,k} g_{j,k}^{(2)}f_j(v_{j}(x-1))f_k(v_{k}(x)) \notag\\
%&+&\sum\limits_{n\ge 2}^N\sum\limits_{\mu_{n-1}}\sum\limits_k g_{\mu_{n-1},k}^{(n)}\tilde{v}^{(n-1)}_{\mu_{n-1}}(x-1)f_k(v_{k}(x)) , 
%\end{eqnarray}
\section{the retokenization group}\label{Sec:model}

The following Hamiltonian 
\begin{eqnarray}\label{Eqn:Ising2_tilde}
H_2(x) = \sum\limits_{j,k} g_{j,k}^{(2)}v_{j}(x-1)v_{k}(x) 
\end{eqnarray}
describes a chain of $d$-dimensional sub-networks (with firing rate vector $v(x)$), connected by translationally-invariant synapses $g^{(2)}(x)=g^{(2)}(x+1)$ (discussed below). (We define energy $E=-H$.) Note that $g_{j,k}^{(2)}\neq g_{k,j}^{(2)}$ in general. We introduce a basis tokenset composed of the $d=26$ characters of the English alphabet, $j,k,l,z\in\{{\bf a},{\bf b},{\bf c},\cdots,{\bf z}\}$ \cite{note_PhonologicalTokens,Boersma2019_deepBoltzmann,Boersma2020_NeuralNets4Phonology}. (Note that we distinguish alphabetic labels from indices via bold font.) Elements $v_j(x)$ encode the overlap of the firing rate vector with the basis token vector $v_j\in\{v_{\bf a},v_{\bf b},\cdots,v_{\bf z}\}$, which we take here to be a one-hot encoding. 

This model bears a resemblance to the synaptic chain seen in songbirds \cite{Long&Jin2010nature_SynapticChainBirdsong}. Information is stored heteroassociatively, meaning that memories are retrieved as the fixed point of the flow of $v(x)$ given $v(x-1)$. We will discuss memory retrieval later, and instead focus on learning.

Learning occurs over time $\tau_d=N_d dt$, which is broken up into length $dt$ time-steps, where $N_d$ is the number of characters in the data string. This data arrives as a sequence of firing states from the input, arriving to $v(x)$ during a $dt$ interval, during which we pin $v(x)$ to the corresponding basis feature vector of that character (or $\vec{0}$, the zero vector, if that character is a space or punctuation). This sequence should be understood as arriving to different sites at different times. For example, at sites $x'\leq x$: ``My name" at $t=0$ pins $v(x)=v_{\bf m}$ and $v(x'<x)=\vec{0}$; then at $t=dt$, $v(x)=v_{\bf y}$, $v(x-1)=v_{\bf m}$, and $v(x'<x-1)=\vec{0}$; then at $t=2dt$, $v(x)=\vec{0}$ because of the space, but $v(x-1)=v_{\bf y}$, etc. This produces the illusion that data is moving down the chain.

For now, we will simply assume the pinning occurs as described above without interference from neighboring neurons in the chain. Learning follows the equation 
\begin{eqnarray}\label{Eqn:g2(t+dt)}
&&g_{j,k}^{(2)}(t+dt,x) = \\
&&(1-\xi_{g})g_{j,k}^{(2)}(t,x) + \xi_gv_j(t,x-1)v_k(t,x) ,  \notag
\end{eqnarray}
having defined $\xi_g\equiv dt/\tau_g$ for time-step $dt$. Note that $\xi_g\rightarrow 0$ is the limit definition of the differential equation $\tau_g\dot{g}_{jk}^{(2)}(x)+g_{jk}^{(2)}(x)=v_j(x-1)v_k(x)$. Correlations in the training data are thus imprinted in an unsupervised way, as a muscle memory, where Hebb's ``fire together wire together" translates into ``practice makes you stronger". 

The effective motion of the data down the chain guarantees $g_{j,k}^{(2)}(x)\simeq g_{j,k}^{(2)}(x+1)$, becoming an equality in the limit $\tau_d/\tau_g\rightarrow 0$ \& $\xi_g\rightarrow 0$. This motivates a simplifying assumption, where we will simply train the synapse at $x$, then copy it to every other synapse at $x'\neq x$. This simplifies training down to a single $x$-independent $g^{(2)}(t)$. We therefore need only focus on the state of the pair \{$v(x-1,t)$,$v(x,t)$\} at fixed $x$. Because the data flows at a rate of one token per time-step, the pair behaves as a 2-token window which observes all the bigrams in the data string.

Once the final bigram in the data string is reached, we stop training $g^{(2)}_{jk}$, and begin growing additional features corresponding to learned bigrams. Note that the tensor product $v_{j}v_{k}$ can be understood as representing a single vector in a $d^2$-dimensional space of bigram features, with combined index $(jk)\in\{{\bf aa},{\bf ab},\cdots,{\bf ba},{\bf bb},\cdots,{\bf zz}\}$. Instead of keeping all $d^2$-many bigrams, we define a new set of vectors $\tilde{v}_{\mu_2}^{(2)}$ with index $\mu_2$, whose dimension $d_2\equiv\dim(\mu_2)<d^2$ is equal to the number of $(jk)$-bigrams for which $g^{(2)}_{jk}>\epsilon_2$. This defines for us a projection map between the $d^2$ and $d_2$ spaces, 
\begin{eqnarray}
P_2(x) = \sum_{\mu_2}\sum_{j,k} P_2^{\mu_2,j,k}\tilde{v}_{\mu_2}^{(2)}(x-1)v_{j}(x-1)v_{k}(x) , 
\end{eqnarray}
where $P_2^{\mu_2,j,k}=1$ if $g_{jk}^{(2)}>\epsilon_2$ (else $=0$). The tensor $P_2^{\mu_2,j,k}$ maps $v_{j}v_{k}\rightarrow \tilde{v}_{\mu_2}^{(2)}$ if $(jk)$ is a relevant bigram, or maps to $0$ otherwise. We choose a convention for the $x$-index $\tilde{v}(x-1)$ to correspond to that of the left-most token of the projected tokens. For example, if the only relevant bigrams are $\mu_2\in\{{\bf ab},{\bf ba}\}$, then $P_2(x)=\tilde{v}_{\bf ab}^{(2)}(x-1)v_{\bf a}(x-1)v_{\bf b}(x)+\tilde{v}_{\bf ba}^{(2)}(x-1)v_{\bf b}(x-1)v_{\bf a}(x)$, where $\tilde{v}_{\bf ab}=(1,0)$ \& $\tilde{v}_{\bf ba}=(0,1)$.

We have in effect merged the 2-point terms into a single compressed vector, which we use to define another 2-point Hamiltonian 
\begin{eqnarray}
H_3(x) = \sum_{\mu_2}\sum_k g^{(3)}_{\mu_2,k}\tilde{v}_{\mu_2}^{(2)}(x-2)v_k(x) . 
\end{eqnarray}
Note we can do the following: $g^{(3)}_{\mu_2,k}\tilde{v}_{\mu_2}^{(2)}v_k = \sum_{l,j}g^{(3)}_{\mu_2,k}P_2^{\mu_2,l,j}v_lv_jv_k$, where $x$-dependence can be inferred by the index ordering. We then define $g^{(3)}_{l,j,k}=\sum_{\mu_2}P_2^{\mu_2,l,j}g^{(3)}_{\mu_2,k}$ as the 3-index representation of $g^{(3)}$. Thus it can be seen that 
\begin{eqnarray}
H_3(x) = \sum_{ljk} g^{(3)}_{l,j,k}v_l(x-2)v_j(x-1)v_k(x) . 
\end{eqnarray}
encodes information about trigrams. Note $g^{(3)}_{l,j,k}$ is defined w.r.t $g^{(3)}_{\mu_2,k}$. The latter is learned following 
\begin{eqnarray}\label{Eqn:g3(t+dt)}
&&g_{\mu_2,k}^{(3)}(t+dt) = \\
&&(1-\xi_{g})g_{\mu_2,k}^{(3)}(t) + \xi_g\tilde{v}_{\mu_2}^{(2)}(x-2,t)v_k(x,t) ,  \notag
\end{eqnarray}
where our new learning window is the pair \{$\tilde{v}_{\mu_2}^{(2)}(x-2,t)$,$v_k(x,t)$\}.

In the language of physics theory, such change of representations are called ``gauge" transformations. These gauge transformations are not physical symmetries, but are a redundancy in the description of the physical system. For example, if {\bf cat} is a learned word, then it represents the fact that $\tilde{v}_{\bf cat}^{(3)}$, $\tilde{v}_{\bf ca}^{(2)}v_{\bf t}$, $v_{\bf c}\tilde{v}_{\bf at}^{(2)}$, \& $v_{\bf c}v_{\bf a}v_{\bf t}$ are all different representations of the same learned token, e.g $\tilde{v}_{\bf cat}^{(3)}=P_3\big(v_{\bf c}P_2\big(v_{\bf a}v_{\bf t}\big)\big)$. Because of this, we'll call such projections/reprojections {\it retokenization}. Non-learned tokens vanish under retokenization. We call tokens which don't vanish under retokenization {\it smooth}. (More on this later.)

We continue this procedure layer-by-layer. This derives the hierarchical-chain Hamiltonian (Fig \ref{Fig:hierarchicalchain}), 
\begin{eqnarray}\label{Eqn:IsingN_tilde}
H(x) = \sum_{n\geq 2}\sum_{\mu_{n-1}}g^{(n)}_{\mu_{n-1},k}\tilde{v}_{\mu_{n-1}}^{(n-1)}(x-n+1)v_k(x) , 
\end{eqnarray}
where we introduce $\mu_1$ as an additional index of the basis tokenset, i.e $d=\dim(\mu_1)=\dim(k)$; and thus $\tilde{v}^{(1)}(x)\equiv v(x)$. Learning follows the Hebbian update rule 
\begin{eqnarray}\label{Eqn:g(t+dt)}
&&g_{\mu_{n-1},k}^{(n)}(t+dt) = \\
&&(1-\xi_{g})g_{\mu_{n-1},k}^{(n)}(t) + \xi_g\tilde{v}_{\mu_{n-1}}^{(n-1)}(x-n+1,t)v_k(x,t) ,  \notag
\end{eqnarray}
where 
\begin{eqnarray}
\tilde{v}_{\mu_{n-1}}^{(n-1)}(x-n+1)&=& \sum_{\mu_{n-2}\cdots \mu_2}\sum_{z\cdots lj}P_{n-1}^{\mu_{n-1},\mu_{n-2},z}\cdots P_2^{\mu_2,lj} \notag\\
&&\times v_{z}(x-n+1)\cdots v_{l}(x-2)v_{j}(x-1) \notag
\end{eqnarray}
is the projection of the trailing context tokens. The projectors have the explicit form 
\begin{eqnarray}
&&P_n(x) = \sum_{\mu_n\mu_{n-1}k}P_n^{\mu_n,\mu_{n-1},k} \\
&& \times\tilde{v}_{\mu_n}^{(n)}(x-n+1)\tilde{v}_{\mu_{n-1}}^{(n-1)}(x-n+1)v_k(x) \notag ,  
\end{eqnarray}
where $P_n^{\mu_n,\mu_{n-1},k}=1$ if $g^{(n)}_{\mu_{n-1},k}>\epsilon_n$ (else $=0$). Note that $H_n(x)$ has an equivalent left-tokenized form, 
\begin{eqnarray}\label{Eqn:IsingN_tilde_left}
&&\mathcal{L}_n\big(H_n(x+n-1)\big) = \\
&&\sum_{\mu_{n-1}}g^{(n)}_{k,\mu_{n-1}}v_k(x)\tilde{v}_{\mu_{n-1}}^{(n-1)}(x+1) , \notag
\end{eqnarray}
where $\mathcal{L}_n$ is constructed in appendix. It can be understood as taking $g_{\mu_{n-1},k}^{(n)}\rightarrow g_{k,\mu_{n-1}}^{(n)}\neq g_{\mu_{n-1},k}^{(n)T}$. (In our notation, indices should only be manipulated via projector maps.) We consider $\mathcal{L}(H)$ as acting $\mathcal{L}_n$ upon every $H_n$. Form $H$ (or $\mathcal{L}(H)$) makes explicit the right-most (or left-most) $v$-token in the chain. 

Crucially, we demand our $g^{(n)}$ be smooth. This guarantees that all learned $n$-grams (i.e elements of $\mu_n$) be composed only of $(n-1)$-grams (elements of $\mu_{n-1}$). This is guaranteed if $g^{(n)}(x)=g^{(n)}(x+1)$ and we demand that $H_n(x)$ be invariant under applying $\mathcal{L}_n$ followed by $\mathcal{L}_n^{-1}$. We smooth every $H_n$ after learning and before constructing $\tilde{v}^{(n)}$. This is the same as projecting $H$ onto the smooth tokenset, which is a low-energy translationally-invariant subspace. 

To understand this step, let's continue our earlier example with $\mu_2\in\{{\bf ab},{\bf ba}\}$. If the training data strings contains segment $[v_{\bf a},v_{\bf b},v_{\bf b},v_{\bf a}]$, naive calculation of Eqn \ref{Eqn:g(t+dt)} for $n=3$ would lead to $g^{(3)}_{\bf abb},g^{(3)}_{\bf bba}>0$. However, {\bf bb} is not a learned bigram (i.e $g_{\bf bb}^{(2)}\leq\epsilon_2$ s.t $P_2(v_bv_b)=0$). The smoothness constraint demands $g^{(3)}_{\bf abb}=g^{(3)}_{\bf bba}=0$. There are two possible remedies: (1) modify the update rule Eqn \ref{Eqn:g(t+dt)} to first check if $(l,j,k)$ is a smooth trigram before adding its contribution to $g^{(3)}$. If $(l,j)$ is smooth, then its sufficient to check if $(j,k)$ is smooth. (This trick generalizes to higher $n$.) Or (2), post-learning $g^{(3)}$, apply the smoothing operation $\mathcal{L}_3^{-1}\mathcal{L}_3$ as described previously.

The constraint of smoothness is an essential assumption of our model. We make this assumption because it guarantees that speech is broken up into words, and further imbues the model with a natural morphology -- Sec \ref{Sec:ralang}. An early glimpse of this can be seen if we extend our example by arguing that $\epsilon_3$ is such that only $g_{\bf aba}^{(3)}>\epsilon_3$ is a learned feature at $n=3$. The smooth tokenset is thus $\{\tilde{v}^{(2)}_{\bf ab},\tilde{v}^{(2)}_{\bf ba},\tilde{v}^{(3)}_{\bf aba}\}$. As consequence, it is impossible for the model to grow additional layers at $n>3$, because no 4-gram exists which is composed only of the smooth set. See Fig \ref{Fig:AliceHierarchy}-b.

We understand the group structure of retokenization as arising because we are using a dense hierarchical Hopfield network \cite{DenseHierarchyHopfield_Krotov2021} to model a sparse graph (the smooth tokenset). As stated previously $v_{\bf c}v_{\bf a}v_{\bf t}$ \& $\tilde{v}_{\bf cat}^{(3)}$ are redundant representations of the same smooth token. Their energies are identically $g^{(3)}_{\bf cat}$, which is guaranteed by construction (a gauge symmetry), and not because of some symmetry of the correlations learned from the data. This is fundamentally unlike tokenization schemes where both letters, words, and subwords are elements of the same vector space \cite{Becker&Kahn2024_NotesMathGPT,Gage1994_BPE,Schuster&Nakajima2012_SubwordTokenization}.

\subsection{next-token prediction (a.k.a inference)}

Note that Eqn \ref{Eqn:IsingN_tilde} is equivalent to 
\begin{eqnarray}\label{Eqn:IsingN}
&&H = \sum\limits_{x}\sum\limits_{j} v_j(x)\Bigg(\sum\limits_{k}g^{(2)}_{jk}v_k(x+1) \\
&&+\sum\limits_{kl}g^{(3)}_{jk}v_k(x+1)v_l(x+2) + \cdots \notag\\
&&+ \sum_{kl\cdots z}g_{jkl\cdots z}^{(N)}v_k(x+1)v_l(x+2)\cdots v_z(x+N-1) \Bigg) \notag
\end{eqnarray}
under retokenization. This is an $N$-point Ising model. It perscribes an energy landscape dictated by the correlations $g_{jk\cdots l}^{(n)}$ between the product of $n$ vectors $v_j(x)$ at sites $x$ in a string. The size of the string is set by the number of context tokens $[v(1),v(2),\cdots,v(N-1)]$, which act as a boundary condition for the next token $v(N)$. It is a type of n-gram model. Inference follows from measuring $v(N)$, which is a superposition equal to the gradient $v(N)=f\big({\partial H}/{\partial v(N)}\big)$, where $f$ is any bounded function. (Discussed below.) This is best done by taking the derivative w.r.t the right-tokenized Eqn \ref{Eqn:IsingN_tilde} (first summing $H=\sum_xH(x)$):
\begin{eqnarray}\label{Eqn:partialH}
\frac{\partial H}{\partial v_k(N)} = \sum_{n=2}g_{\mu_{n-1},k}^{(n)}\tilde{v}_{\mu_{n-1}}^{(n-1)}(N-n+1) .  
\end{eqnarray}
If instead $v(1)$ is free, then we can use the left-tokenized Eqn \ref{Eqn:IsingN_tilde_left} to do left inference.

In order to collapse the superposition into meaningful output, we introduce a ``measurement", which samples from a probability distribution $\rho_k(v_k(N)|\vec{v}_{\text{context}})$. A theory for how measurement arises biologically is outside the scope of this work, but one option is to define 
\begin{eqnarray}\label{Eqn:rho}
\rho_k(v_k(N)|\vec{v}_{\text{context}}) = \frac{v_k(N)}{\sum_jv_j(N)} . 
\end{eqnarray}
Lastly, it is possible to bias inference toward preferring longer correlations by scaling $g_{\mu_{n-1},k}^{(n)}\rightarrow \beta^n g_{\mu_{n-1},k}^{(n)}$ for $\beta\gg 1$. Such a factor is a hyperparameter akin to the inverse-temperature of a maximum entropy distribution.

Note that more properly, $u_k(t)$ is governed by $\tau_v\dot{u}_k+u_k={\partial H}/{\partial v_k}$. The firing rate $v_k$ is a function of $u_k$, a.k.a the activation function $v_k = f(u_k)$. In this work, we assume a maximum firing rate $f(u_k)\leq\Lambda_v$ (and choose units s.t $\Lambda_v=1$). For our purposes, we do not care about the firing rate curve, only the state of firing.

Notably, it is possible to generate $N$-point correlations of Eqn \ref{Eqn:IsingN} type, by assuming the existence of fast equilibrated hidden neurons with non-linear activation functions \cite{HopfieldNeurobiology_Krotov&Hopfield2021}. As mentioned in the intro, we find that this assumption does not lead to a scientific explanation for the data. It's possible a reason for this lies in the assumptions about the timescales of neurons which estimate stimuli, such as the cosine turning curve of the cricket \cite{Salinas&Abbott1994_CosineTuningCurve}. Sensory and motor processing is noticeably slower than language. Thus we work in the limit of fast cognitive processing, where only the state of firing is necessary, and not the firing rate curve.

We will return to a general discussion of how the model scales during training in Sec \ref{Sec:explodingdimensions}. In the next section, we will discuss the properties of the hierarchal tokenset with the aid of a minimal analytical example.

\subsection{an example}\label{Sec:retokenization}

Consider the following example string: ``I run, he runs, they ran." Learning every correlation in this string is equivalent to learning the tokenset: $\tilde{v}_{\bf ru}^{(2)}$,$\tilde{v}_{\bf un}^{(2)}$,$\tilde{v}_{\bf ra}^{(2)}$,$\tilde{v}_{\bf an}^{(2)}$,$\tilde{v}_{\bf th}^{(2)}$,$\tilde{v}_{\bf he}^{(2)}$,$\tilde{v}_{\bf ey}^{(2)}$; $\tilde{v}_{\bf run}^{(3)}$,$\tilde{v}_{\bf ran}^{(3)}$,$\tilde{v}_{\bf the}^{(3)}$,$\tilde{v}_{\bf hey}^{(3)}$; $\tilde{v}_{\bf they}^{(4)}$ -- see Fig \ref{Fig:featuremaps}. This tokenset defines a series of projector maps (where we suppress $x$ and treat vector products as non-commuting): $P_2 = \tilde{v}_{\bf ru}^{(2)}v_{\bf r}v_{\bf u}+\tilde{v}_{\bf un}^{(2)}v_{\bf u}v_{\bf n}+\tilde{v}_{\bf ra}^{(2)}v_{\bf r}v_{\bf a}+\tilde{v}_{\bf ey}^{(2)}v_{\bf e}v_{\bf y}$; $P_3=\tilde{v}_{\bf run}^{(3)}\tilde{v}_{\bf ru}^{(2)}v_{\bf n}+\tilde{v}_{\bf ran}^{(3)}\tilde{v}_{\bf ra}^{(2)}v_{\bf n}+\tilde{v}_{\bf the}^{(3)}\tilde{v}_{\bf th}^{(2)}v_{\bf e}+\tilde{v}_{\bf hey}^{(3)}\tilde{v}_{\bf he}^{(2)}v_{\bf y}$; $P_4=\tilde{v}_{\bf they}^{(4)}\tilde{v}_{\bf the}^{(3)}v_{\bf y}$.

Consider the token $\tilde{v}_{\bf run}^{(3)} = P_3(P_2(v_{\bf r}v_{\bf u})v_{\bf n})$. If we were to try to grow the product, say by $v_{\bf h}$, we would find $P_4(\tilde{v}_{\bf run}^{(3)}v_{\bf h})=0$. The reason for this is multiple: not only is $v_{\bf r}v_{\bf u}v_{\bf n}v_{\bf h}$ not a correlation in the training example, but neither is $v_{\bf n}v_{\bf h}$ nor $v_{\bf u}v_{\bf n}v_{\bf h}$. 

The string $v_{\bf r}v_{\bf u}v_{\bf n}v_{\bf h}v_{\bf e}$ can be tokenized as $\tilde{v}_{\bf run}^{(3)}\tilde{v}_{\bf he}^{(2)}$ by starting anywhere in the string where $P_2$ doesn't vanish, and growing the token left/right until its boundaries have been reached. Note that ``runh" is not a pattern used in any commonly written English word \cite{wordfinder}.

\begin{figure}
    \centering
    \includegraphics[width=1.0\linewidth]{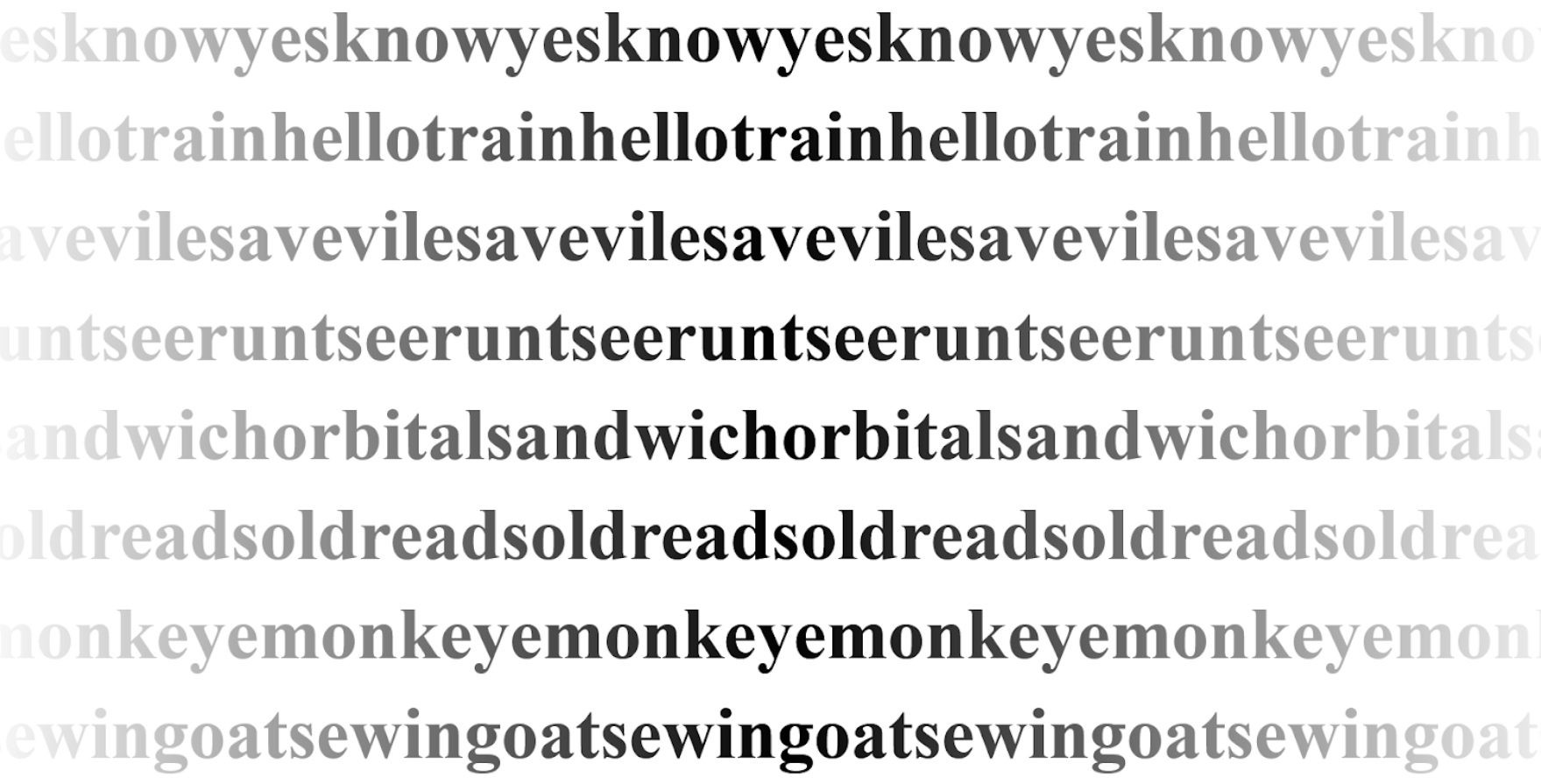}
    \caption{Infinite strings composed of two repeating words. Boundary information is hidden to prevent the reader from using it to tokenize. The reader can tokenize by looking for the non-word-forming patterns which contain the word boundaries. Only ``savevile" and ``soldread" do not tokenize into 2 words uniquely.}
    \label{Fig:savevile}
\end{figure}

Once learned, neighboring words can be discerned by the irregular patterns which form at their boundaries. This property is not special to the given example, but is a general property of written English. A skeptical reader can explore this themselves by combining word pairs and asking if all the overlaps are elements of the ruleset for constructing words. For cases like ``savevile" (save+vile), which has recognizable ``evil" at its boundary -- see Fig \ref{Fig:savevile} -- a unique tokenization can still eventually be found with the aid of the left/right edge, as well as other patterns (e.g ``avev"). True violations do exists, but lend themself to an illusion of multiplicity; for example ``$\text{petsmart}$" parsed as either pets+mart or pet+smart.

\section{exploding dimensions and forgetting}\label{Sec:explodingdimensions}

In the previous section, we examined tokenization with a minimal example. Here, we analyze model scaling on a dataset of token length $N_d$. Since the model is trained hierarchically from $n=2$, we must determine how the effective dimension $d_n\equiv\text{dim}(\mu_n)$ grows with $n$. Three factors influence $d_n$: (1) $d_{n-1}$, the previous scale’s token count; (2) $\epsilon_n$, the cutoff; and (3) $\tau_g$, the synaptic decay time, which we redefine as $\tau_g = N_g dt$, where $N_g$ is roughly the number of time steps before information is rapidly forgotten (see appendix). Thus $N_g\ge N_d$ is required in order to retain all information. Here, we assume $N_g\rightarrow\infty$ to study learning without forgetting.

%In the previous section, we explored tokenization using a minimal training example. Here we explore how the model scales on a more general dataset of token length $N_d$. Since the model is fit hierarchically, starting from $n=2$, we need to know how the effective dimension $d_n\equiv\text{dim}(\mu_n)$ scales with increasing $n$. There are three factors which determine the $d_n$: (1) $d_{n-1}$, the number of tokens at the previous scale; (2) $\epsilon_n$, the cutoff; and (3) $\tau_g$, the decay time of the synapses. The latter can be redefined as $\tau_g = N_g dt$, where $N_g$ is the number of time-steps before information imprinted into the synapses is forgotten (see appendix). Thus in order for the synapses to retain all the information read from a string, it is necessary to choose $N_g\ge N_d$. For now, we will assume $N_g\rightarrow\infty$ and study the case of learning absent forgetting.

\begin{figure}
    \centering
    \includegraphics[width=1.\linewidth]{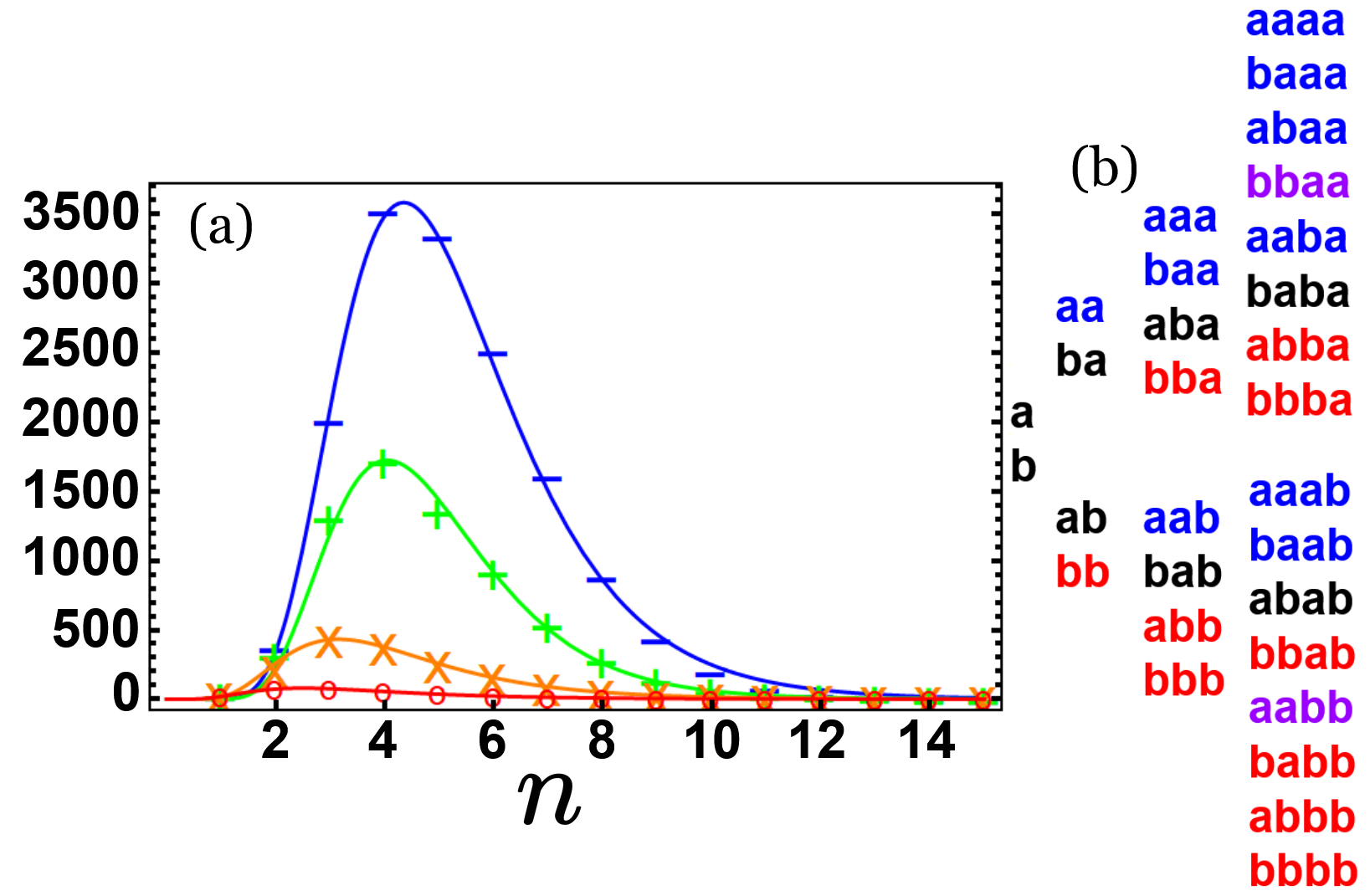}
    \caption{(a) Hierarchy of unique $n$-grams from text taken from {\it Alice in Wonderland}. Different curves correspond to increasing text sizes $N_d\in\{235,2336,22762,107777\}$, with $N_d=107777$ being the completed text. Solid lines show fits to log-norm (see appendix): $F(n,1.22,.55,320)$, $F(n,1.35,.45,1700)$, $F(n,1.52,.35,6500)$, \& $F(n,1.6,.36,15000)$ resp. (b) Hierarchy of language with $d=2$. Colors show how constraints on earlier levels, ${\color{blue} g^{(2)}_{\bf aa}=0}$ \& ${\color{red} g^{(2)}_{\bf bb}=0}$ \& ${\color{purple} g^{(2)}_{\bf aa}g^{(2)}_{\bf bb}=0}$, limit the allowed growths at later levels. The hierarchy is stable if both {\bf aa} \& {\bf bb} are disallowed, but collapses if more (say $g_{\bf aba}^{(3)}=0$) is introduced.}
    \label{Fig:AliceHierarchy}
\end{figure}

\begin{figure}
    \centering
    \includegraphics[width=.85\linewidth]{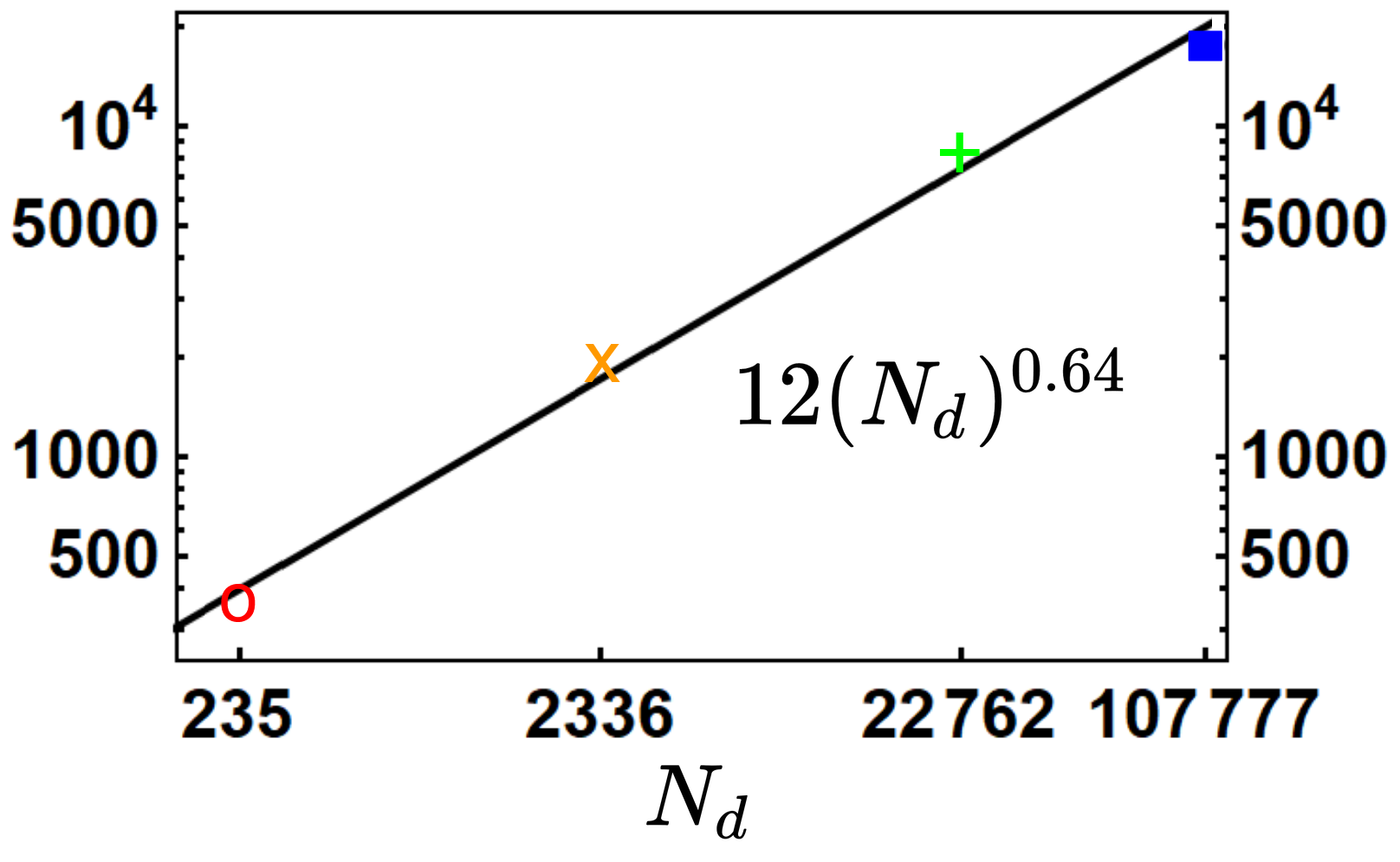}
    \caption{The total number of $n$-grams, $\sum_n d_n$, as a function of text size.}
    \label{Fig:AliceNgramCount}
\end{figure}

When $\epsilon_n=0$, the model learns all unique $n$-grams of length $n$ in the text. Fig \ref{Fig:AliceHierarchy} shows the number of unique intra-word $n$-grams ($d_n$) versus $n$ and text size ($N_d$) for Alice in Wonderland \cite{GutenbergAlice}. The curve peaks at $3\leq n_{\text{peak}}\leq 4$ before collapsing. This distribution has been estimated as log-normal \cite{Herdan1958_logNormal_intraword,Williams1940_logNormal_interwords,Eckhard&Werner&Markus2001_logNormal_review}. The total $n$-gram count ($\sum_n d_n$, Fig \ref{Fig:AliceNgramCount}) grows polynomial in $N_d$, dominated by the peak. These dominant $n$ create a memory bottleneck, as projectors $P^{(n)}_{\mu_{n},\mu_{n-1},k}$ scale with $d_{n}\times d_{n-1}\times d$, limiting learning without forgetting. Setting $\epsilon_n>0$ mitigates this but prevents full tokenization.

%When $\epsilon_n=0$, the model learns every unique $n$-gram of length $n$ in the text. In Fig \ref{Fig:AliceHierarchy}, we plot the number of unique intra-word n-grams ($d_n$) as a function of $n$ and increasing text size ($N_d$) for the novel {\it Alice in Wonderland} \cite{GutenbergAlice}. For any $N_d$, the shape of the curve with increasing $n$ peaks at some $3\leq n_{\text{peak}}\leq 4$ before collapsing exponentially to zero. (Note $n$ is bounded by the longest word(s) in the corpus.) Previous authors have estimated this distribution as log-normal \cite{Herdan1958_logNormal_intraword,Williams1940_logNormal_interwords,Eckhard&Werner&Markus2001_logNormal_review}. We find the sum number of grams ($\sum_n d_n$ shown in Fig \ref{Fig:AliceNgramCount}) grows polynomial in $N_d$, and is dominated by the peak. These dominate $n$ create a memory bottleneck, because the projectors $P^{(n)}_{\mu_{n},\mu_{n-1},k}$ scale with dimensions $d_{n}\times d_{n-1}\times d$. This places a practical bound on learning without forgetting. Introducing $\epsilon_n>0$ regulates this; but then $d_n$ is no longer equal to the total number of $n$-grams, and the text cannot be completely tokenized.

Note that short-term human memory is demonstrably limited by string length more than decay time \cite{Shiffrin1970_forgettingLongStrings,Gersham&Fiete&Irie2025_KeyValueMemory}. A handful of random words can be remembered for a time, but long enough strings ($\sim 5-20$ words) are forgotten \cite{Shiffrin1970_forgettingLongStrings,Gersham&Fiete&Irie2025_KeyValueMemory}. Here we find that hierarchical learning is similarly limited by length. This is because a single set of projectors learns all the feature maps used to tokenize the words in the training data. Reintroducing forgetting only makes perfect tokenization impossible. Curiously, these two problems -- the $P_n$ scaling and forgetting -- have the same solution. This is discussed in the next section. Readers interested in hierarchical random language generation can skip onto Sec \ref{Sec:ralang}.

%One solution is to use a sparse representation (e.g COO) for the $P_n$. However, this computational solution provides no additional intuition into the physical problem of computation held by the biology, which still requires training $d_{n}\times d_{n-1}\times d$ possible synaptic connections. Additionally, the time-scale for $\tau_g$ is not really infinite. In the author's experience, details read from a text are lost on some timescale. However, without sufficiently large $\tau_g$, it becomes impossible for our model to learn all the $n$-grams necessary to do general tokenization. Curiously, these two problems -- the $P_n$ scaling and forgetting -- have the same solution. This is discussed in the next section.

\section{Hebbian replay}\label{Sec:replay}

Let us generalize Eqn \ref{Eqn:IsingN_tilde} to include additional terms, 
\begin{eqnarray}\label{Eqn:H_amv}
&&H(x) = \sum\limits_n\sum_{\mu_{n-1},k} g_{\mu_{n-1},k}^{(n)}\tilde{v}_{\mu_{n-1}}^{(n-1)}v_{k} \\
&&+\sum_{\alpha} a_{\alpha}\Big(\sum\limits_n\sum_{\mu_{n},k} m_{\alpha,\mu_n}^{(n)}\tilde{v}_{\mu_n}^{(n)}\Big) + \sum_{\alpha}\psi_{\alpha}a_{\alpha} , \notag
\end{eqnarray}
where we write $\tilde{v}_{\mu_{n-1}}^{(n-1)}(x-n+1)v_k(x)\rightarrow\tilde{v}_{\mu_{n-1}}^{(n-1)}v_k$ with the understanding that $v_k$ is the unpinned final token, with projected trailing context token $\tilde{v}_{\mu_{n-1}}^{(n-1)}$. The vector $a_{\alpha}$ represents the state of added ``auxiliary" neurons, and $\psi_{\alpha}$ is a pinning field (described below). The added term is the most minimal extension of our original model, where we assume a single set of new neurons with simple connections to every layer of the $\tilde{v}_{\mu_n}^{(n)}$-hierarchy through synapses $m_{\alpha,\mu_n}^{(n)}$. Where necessary we will retokenize the 2nd term,
\begin{eqnarray}
\sum\limits_{\mu_n} m_{\alpha,\mu_n}^{(n)}\tilde{v}_{\mu_n}^{(n)} 
= \sum\limits_{\mu_{n-1},k} m_{\alpha,\mu_{n-1},k}^{(n)}\tilde{v}_{\mu_{n-1}}^{(n-1)}v_{k} , 
\end{eqnarray}
where $m_{\alpha,\mu_{n-1},k}^{(n)}=\sum_{\mu_n}m_{\alpha,\mu_n}^{(n)}P^{(n)}_{\mu_n,\mu_{n-1},k}$. This allows us to take derivatives with respect to a final token ($v_k$).

The equations of motion have the general form $\tau_Q\dot{q}+q=\Delta Q$, where $Q=f(q)$, and we've written $\Delta Q\equiv {\partial H}/{\partial Q}$ for short. Here $(q,Q)$ are placeholder representing the dynamical variables. The dynamical variables for neurons are the firing states ($q$) \& rates ($Q$): $v_j=f_j(u_j)$ \& $a_{\alpha}=f_{\alpha}(\mathfrak{a}_{\alpha})$, where $f$ is any bounded function acting on every element of the vector. We treat $f$ as linear for synapse, i.e $Q=q$, so that there is only one dynamical quantity per synapse: $g^{(n)}$ \& $m^{(n)}$.

Note that the flow of the $g_{\mu_{n-1},k}^{(n)}$ is controlled entirely by the state of $v$'s, Eqn \ref{Eqn:g(t+dt)}. The other gradients are 
\begin{eqnarray}\label{Eqn:dv_replay}
&&\Delta v_{k} 
= \sum\limits_n\sum\limits_{\mu_{n-1}}\Big(g_{\mu_{n-1},k}^{(n)}
+ \sum\limits_{\alpha}a_{\alpha}m_{\alpha,\mu_{n-1},k}^{(n)}\Big)\tilde{v}_{\mu_{n-1}}^{(n-1)} \\
&&\Delta a_{\alpha}
= \sum\limits_n\sum\limits_{\mu_{n}} m_{\alpha,\mu_n}^{(n)}\tilde{v}_{\mu_n}^{(n)} + \psi_{\alpha} \\
&&\Delta m_{\alpha,\mu_n}^{(n)}
= a_{\alpha}\tilde{v}_{\mu_n}^{(n)} ,   
\end{eqnarray}
where we take $\tau_m\gg\tau_g\gg\tau_v>\tau_a$, \& $\Lambda_v=\Lambda_a$. Notice that when $m^{(n)}_{\alpha,\mu_n}=0$ and $a_{\alpha}=0$, the original model is returned absent the auxiliaries. In this limit, the flow of $v_{k}$ (and therefore inference) is governed entirely by the information stored in $g_{\mu_{n-1},k}^{(n)}$. If during inference, we were to introduce an auxiliary neuron by forcing it to fire, that neuron would become correlated with the features ($n$-grams) generated during that inference. So long as the flow of $a_{\alpha}$ does not interfere with the inference, which is the case initially when the entries of $m_{\alpha,\mu_n}^{(n)}$ are small, then the synapses of the auxiliary neurons learn word embeddings. 

To make this more concrete, let's define replay as a series of replay cycles. Each cycle is a game of generating a string of tokens through inference, starting with some randomly sampled initial token. We end the inference once the product of the context tokens with the measured value of the next token has a null projection, i.e is non-smooth. Left inference is first performed, which finds the left word boundary. Multiple cycles can then be performed as right-inference restarting from the left boundary. Throughout a cycle, we choose a random auxiliary neuron to be pinned high, e.g $\psi_{\alpha}=\Lambda_{\psi}\delta_{\alpha,1}$ where $\Lambda_{\psi}\gg\Lambda_a$. The effect of which is that the replayed $n$-grams are imprinted in the synaptic connections with this neuron. The $a_1$ thus becomes an embedding tied to these $n$-grams. 
 
%By retokenizing and performing left inference, the left boundary can be found, allowing replay of all the features of a single word (see Supplement).

Taking the limit $\tau_a/\tau_v\rightarrow 0$, we can derive a formula for next-token inference due to these embeddings, 
\begin{eqnarray}\label{Eqn:vvmv}
&&\Delta v_{k} = \\
&& \sum\limits_{\alpha}f_{\alpha}\Big(\sum\limits_{n'}\sum\limits_{\mu_{n'}'}\tilde{v}_{\mu_{n'}'}^{(n')}m_{\alpha,\mu_{n'}'}^{(n')}\Big)\sum\limits_{n}\sum\limits_{\mu_{n-1}}m_{\alpha,\mu_{n-1},k}^{(n)}\tilde{v}_{\mu_{n-1}}^{(n-1)} ,  \notag
\end{eqnarray}
where the $\tilde{v}$ are the projections of the trailing context tokens. Notice that the first $\tilde{v}_{\mu_n}^{(n)}$ selects the embedding $\alpha$; where then the second $\tilde{v}_{\mu_{n-1}}^{(n-1)}$ informs the flow for the final infered token from the $n-1$ tokens behind it. This has the effect that even a single $n$-gram, projected out of the context, can trigger the embedding. Thus allowing the model to infer an entire word from partial information. If the context $n$-grams are features of multiple different embeddings, then $a_{\alpha}$ flows to a fixed point which is a superposition. A tie can be broken by additional context. %Such a behaviour could provide an explanation for errors in human speech where a wrong, but similar sounding, word is produced.

Crucially, this learning of $m_{\alpha,\mu_n}^{(n)}$ frees up $g_{\mu_{n-1},k}^{(n)}$ to learn additional features from the data. This makes possible continual learning without forgetting. Additionally, information can be retained more efficiently (discussed below) in the higher layers of the network. 
%Because $\tau_m\gg\tau_g$, the information stored in the $g_{\mu_{n-1},k}^{(n)}$ are eventually forgotten, nevertheless, its information can be recalled because of the embeddings stored now in $m_{\alpha,\mu_n}^{(n)}$. 

\subsection{compression}

Learning the word embeddings in $a_{\alpha}$ allows for a compression of the synapses. First notice that 
\begin{eqnarray}\label{Eqn:mv_uncompressed}
&&\sum_{\mu_n}m_{\alpha,\mu_n}^{(n)}\tilde{v}_{\mu_n}^{(n)} = \\
&&\sum_{\mu_{n},\cdots,\mu_2}\sum_{jk\cdots z}m_{\alpha,\mu_n}^{(n)}P_{n}^{\mu_n,\mu_{n-1},z}\cdots P_{2}^{\mu_2,j,k}\Big(v_{j}v_{k}\cdots v_{l}v_{z}\Big) \notag
\end{eqnarray}
tells us that the auxiliary synapses learn how to project up from the basis set. For fixed $\alpha$, we can interpret Eqn \ref{Eqn:mv_uncompressed} as an operator string capped by the vector $m_{\mu_n}(\alpha)\equiv m_{\alpha,\mu_n}^{(n)}$. For convenience, we'll further redefine this vector to be a matrix row, $m_{\mu_n}(\alpha)\equiv m_{1,\mu_n}(\alpha)$, and perform an SVD ($m=UDV$) decomposition: $m_{1,\mu_n}(\alpha)=\sum_{\beta_n^{\alpha}}\hat{m}_{1,\beta_n^{\alpha}}V_{\beta_n^{\alpha},\mu_n}$, where $\hat{m}=UD$. Note that there are unique indices $\beta_n^{\alpha}$ per $n$ \& $\alpha$.

Contract $V$ with $P_n$, $\tilde{P}_{n}^{\beta_n^{\alpha},\mu_{n-1},z}=\sum_{\mu_n}V_{\beta_n^{\alpha},\mu_n}P_n^{\mu_n,\mu_{n-1},z}$. The $\dim(\beta_n^{\alpha})<\dim(\mu_n)$ since it only carries the $n$-grams relevant to $\alpha$. Thus $\tilde{P}_n$ has been compressed. Next we merge the right two indices of the projector, which we simply write as $(\mu_{n-1}z)$, and again perform an SVD (or QR) decomposition: $\tilde{P}_n^{\beta_n^{\alpha},(\mu_{n-1}z)}=\sum_{\omega}U_{\beta_n^{\alpha},\omega}V_{\omega,(\mu_{n-1}z)}$. Both $U$ \& $V$ are unitary since the projector's non-zero eigenvalues are $1$. (Note we suppress the $n$ \& $\alpha$ dependence of temporary indices, like $\omega$, which show up at intermediate steps but not the final form.)

Then contract $T_{\omega,z,\mu_{n-2},l}=\sum_{\mu_{n-1}}V_{\omega,\mu_{n-1},z}P_{n-1}^{\mu_{n-1},\mu_{n-2},l}$. Merging the left and right two indices allows for another decomposition $T_{(\omega z),(\mu_{n-2} l)} = \sum_{\beta_{n-1}^{\alpha}}U'_{(\omega z),\beta_{n-1}^{\alpha}}D'_{\beta_{n-1}^{\alpha}}V'_{\beta_{n-1}^{\alpha},(\mu_{n-2} l)}$. We then define $\hat{P}_{n}^{\beta_n^{\alpha},\beta_{n-1}^{\alpha},z}=\sum_{\omega}U_{\beta_n^{\alpha},\omega}U'_{\omega,z,\beta_{n-1}^{\alpha}}D'_{\beta_{n-1}^{\alpha}}$, and $\tilde{P}_{n-1}^{\beta_{n-1}^{\alpha},\mu_{n-2},l}=V'_{\beta_{n-1}^{\alpha},\mu_{n-2},l}$. Projector $\hat{P}_n$ has been fully compressed.

Repeat this process until the final projector (at $n=2$) is compressed. The effect is that we have a new string of projectors, where we have replaced $\mu_n\rightarrow\beta_n^{\alpha}$ where $\beta_n^{\alpha}$ carries only the information relevant to $\alpha$. The right hand side of Eqn \ref{Eqn:mv_uncompressed} becomes 
\begin{eqnarray}\label{Eqn:mv_compressed}
= \sum_{\beta_n^{\alpha},\cdots,\beta_2^{\alpha}}\sum_{jk\cdots z}\hat{m}_{\beta_n^{\alpha}}^{(n)}\hat{P}_{n}^{\beta_n^{\alpha},\beta_{n-1}^{\alpha},z}\cdots \hat{P}_{2}^{\beta_2^{\alpha},j,k} 
\Big(v_{j}v_{k}\cdots v_{l}v_{z}\Big) . \notag\\
\end{eqnarray} 

The reasoning for this dance of contractions \& decompositions is in the tensor product structure of the projectors. Such products have a large gauge freedom \cite{mps_Schollwock2011,mps_Parker&Cao&Zaletel2020}, arising from the fact that the individual tensors generically carry additional information not necessary for computing their product. Consider (from the minimal example of Sec \ref{Sec:retokenization}) the tensor $\tilde{v}_{\bf they}^TP_4P_3P_2$. It acts on a string of four basis tokens, and equals $1$ ($0$) depending on whether that string is $v_{\bf t}v_{\bf h}v_{\bf e}v_{\bf y}$ (else). A generic product of $P_n$ carries this projection map, in addition to all other projections maps. But information about constructing $\tilde{v}_{\bf run}$ is not relevant to constructing $\tilde{v}_{\bf they}$. The game of decompositions/contractions allows information high in the network to be carried down to the lower layers, which prunes $\tilde{v}_{\bf run}$ and other irrelevancies. Thus by learning $m_{\mu_n}(\alpha)$, a collection of disentangled projection maps $\hat{P}_n(\alpha)$ becomes possible.

Disentangling the projection maps guarantees memory scales linearly with the number of words ($d_a$), because each word embedding contributes independently to the total cost. For a single embedding length $N_{\alpha}$, we measure its cost $\gamma_{\alpha}$ as the total number of matrix elements of the projector product string. It is 
\begin{eqnarray}
&&\gamma_{\alpha} = d^2 + \\
&&(d+d^2)\sum_{n=3}^{N_{\alpha}}(N_{\alpha}-n+1) + d\sum_{n=3}^{N_{\alpha}}(n-3)(N_{\alpha}-n+1)^2 , \notag
\end{eqnarray}
or equivalently $\frac{d}{12}\Big(24-46N_{\alpha}+29N_{\alpha}^2-8N_{\alpha}^3+N_{\alpha}^4+6d(4-3N_{\alpha}+N_{\alpha}^2)\Big)$. 

This disentanglement makes inference fully parallelizable, as each $a_{\alpha}$ contributes independently to Eqn \ref{Eqn:vvmv}. Training is also parallelizable -- features can be learned incrementally over smaller steps. For example, By splitting text into $B$ batches of length $N_b$ (e.g. paragraphs), $B$ copies of Eqn \ref{Eqn:H_amv} can generate embeddings simultaneously.

Our compression technique is inspired by Matrix Product States (MPS, a.k.a. TensorTrain) from quantum many-body systems \cite{mps_Schollwock2011,mps_Parker&Cao&Zaletel2020}. Here we leverage the tensor product structure by decomposing along the tokenization direction, which is similar in spirit to MPS methods used for analyzing inter-lengthscale correlations of turbulence structures \cite{mps_Gourianov&all2022_turbulence,mps_Stoudenmire2022_CondMattJournalClub}.

\section{Random languages and the scaling collapse of the hierarchy}\label{Sec:ralang}

%In Sec \ref{Sec:explodingdimensions} we discussed the model scaling during training, with a focus on the limit $\epsilon_n=0$ and $N_g\rightarrow\infty$, such that the model will attempt to learn every $n$-gram in a text length $N_d$. In this limit, the effective dimension $d_n$ is equal to the number of $n$-grams in the text, and the collapse of the curves shown in Fig \ref{Fig:AliceHierarchy} is set entirely by the longest word in the corpus. This demonstrates {\it how} the model learns the hierarchy of correlations already present in the data, but doesn't provide an explanation as to {\it why} such correlations exist in the first place. In this section, we show how our model is not only sufficient to capture such correlations, but is a candidate for their microscopic origin.

In Sec \ref{Sec:explodingdimensions}, we examined model scaling in the limit $\epsilon_n=0$ \& $N_g\rightarrow\infty$, where every $n$-gram in a text of length $N_d$ is learned. Here, $d_n$ is set by the number of $n$-grams in the text, and the curve collapse in Fig \ref{Fig:AliceHierarchy} is determined by the longest word in the corpus. This shows how the model learns hierarchical correlations but not why they exist. In this section, we argue that our model not only captures these correlations but may explain their microscopic origin.

%We do this by growing a random hierarchy (discussed below), then performing replay in order to generate the words stored in that hierarchy. We then study this novel vocabulary, and find it exhibits a distribution of word-forming patterns representative of natural languages. This includes: (1) finite-word length; and a distribution of unique $n$-grams, as a function of length $n$, which peaks at $n_{\text{peak}}$ then collapses (Fig \ref{Fig:peak&collapse}). And (2) a sequence of rank-ordered frequency distributions with similarly-sloped power laws, whose slopes persists for increasing $n$ beyond $n\sim n_{\text{peak}}$ (Fig's \ref{Fig:yjjfgsp}). The similarity and persistence of these slopes is a signature of morphology, i.e a hierarchy where the $n_{\text{peak}}$-grams are the building blocks used to construct the words living at larger $n$. 
%and a lognormal distribution for the number of unique $n$-grams as function of length $n$ (Fig \ref{Fig:peak&collapse}); 

We do this by growing a random hierarchy (discussed below), then perform replay to generate stored words, and analyze the resulting vocabulary. The word-forming patterns of this novel vocabulary mirrors natural languages, exhibiting: (1) finite word length and an $n$-gram distribution that peaks at $n_{\text{peak}}$ before collapsing (Fig 6); and (2) rank-ordered frequency distributions with persistent power-law slopes for $n>n_{\text{peak}}$ (Fig 5b). These stable slopes indicate morphology, where $n_{\text{peak}}$-grams serve as building blocks for longer words.

\begin{figure}
\centering
\begin{subfigure}[t]{0.5\textwidth}
    \centering
    \includegraphics[width=.9\linewidth]{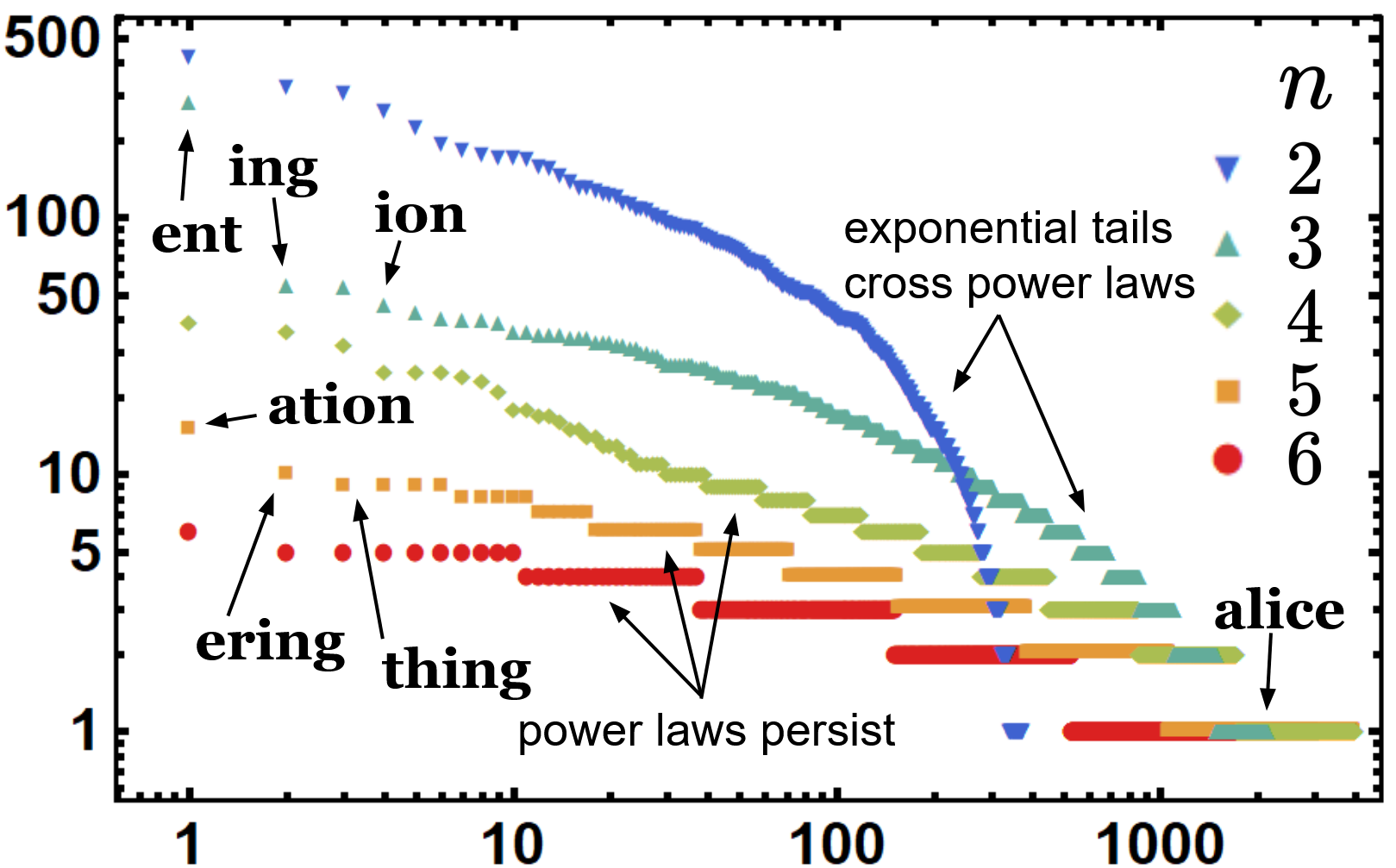}
    \caption{Frequency (vertical axis) of $n$-grams ordered by rank (horizontal axis) for: 2567 unique words from {\it Alice in Wonderland} (full text), with transition to power-law occurring between $n=3-4$.}
    \label{Fig:Alice}
\end{subfigure}\\
%\begin{subfigure}[t]{0.5\textwidth}
%    \centering
%    \includegraphics[width=.9\linewidth]{iiz.png}
%    \caption{1159 words of the random language {\it iiz}, and five example words which contribute to the statisics of the $n$-grams shown. The transition occurs between $n=2-3$. Note $n_{\text{peak}}=3$.}
%    \label{Fig:iiz}
%\end{subfigure}
\begin{subfigure}[t]{0.5\textwidth}
    \centering
    \includegraphics[width=.9\linewidth]{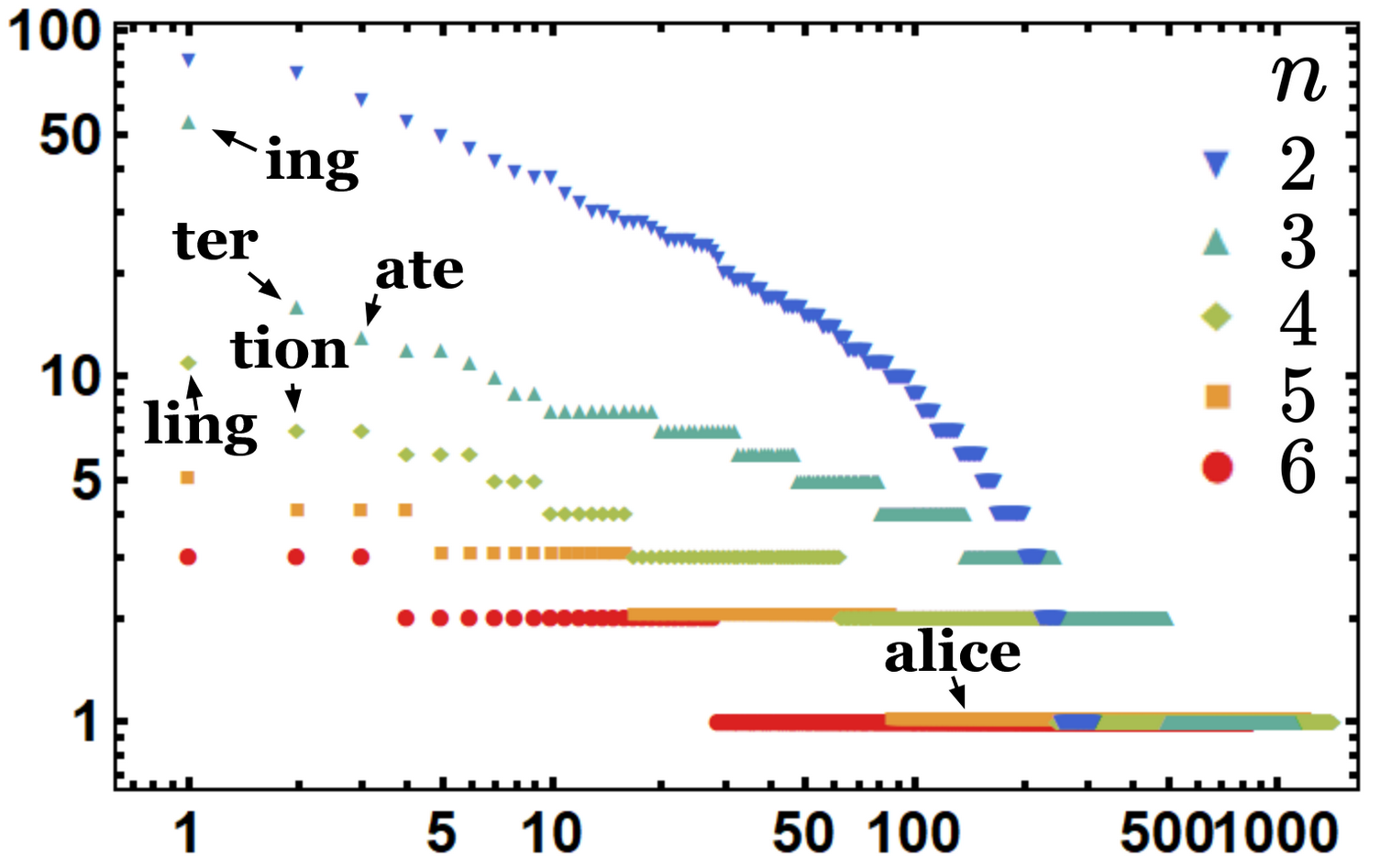}
    \caption{500 words of {\it Alice in Wonderland}. Smaller samples (a few hundred words) are most similar to random language data from growing a hierarchy.}
    \label{Fig:Alice500words}
\end{subfigure}
\begin{subfigure}[t]{0.5\textwidth}
    \centering
    \includegraphics[width=.9\linewidth]{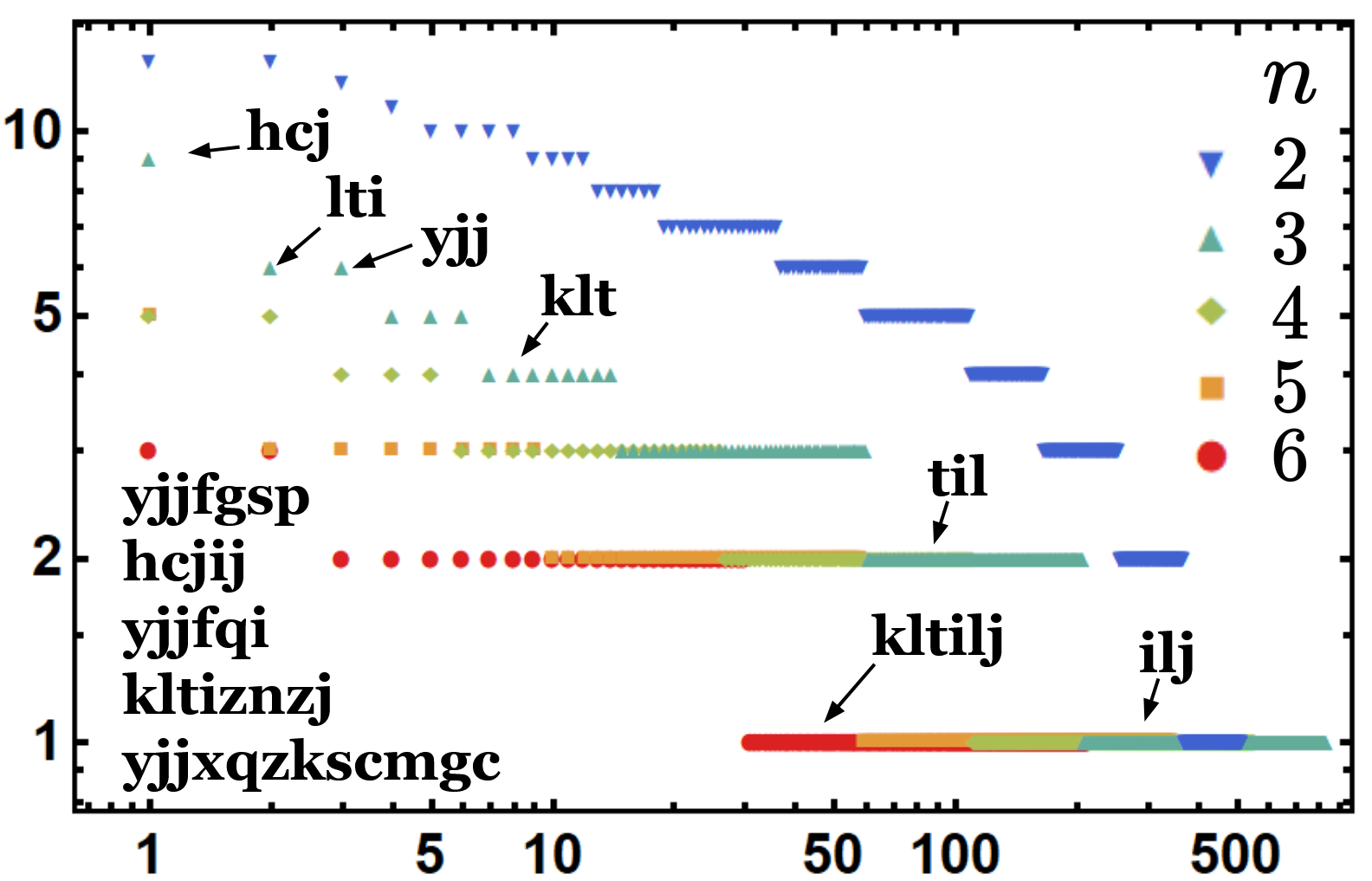}
    \caption{429 words of the random language {\it yjjfgsp}, and five example words which contribute to the statisics of the $n$-grams shown. The transition occurs between $n=2-3$. Note $n_{\text{peak}}=3$.}
    \label{Fig:yjjfgsp}
\end{subfigure}
\begin{subfigure}[t]{0.5\textwidth}
    \centering
    \includegraphics[width=.9\linewidth]{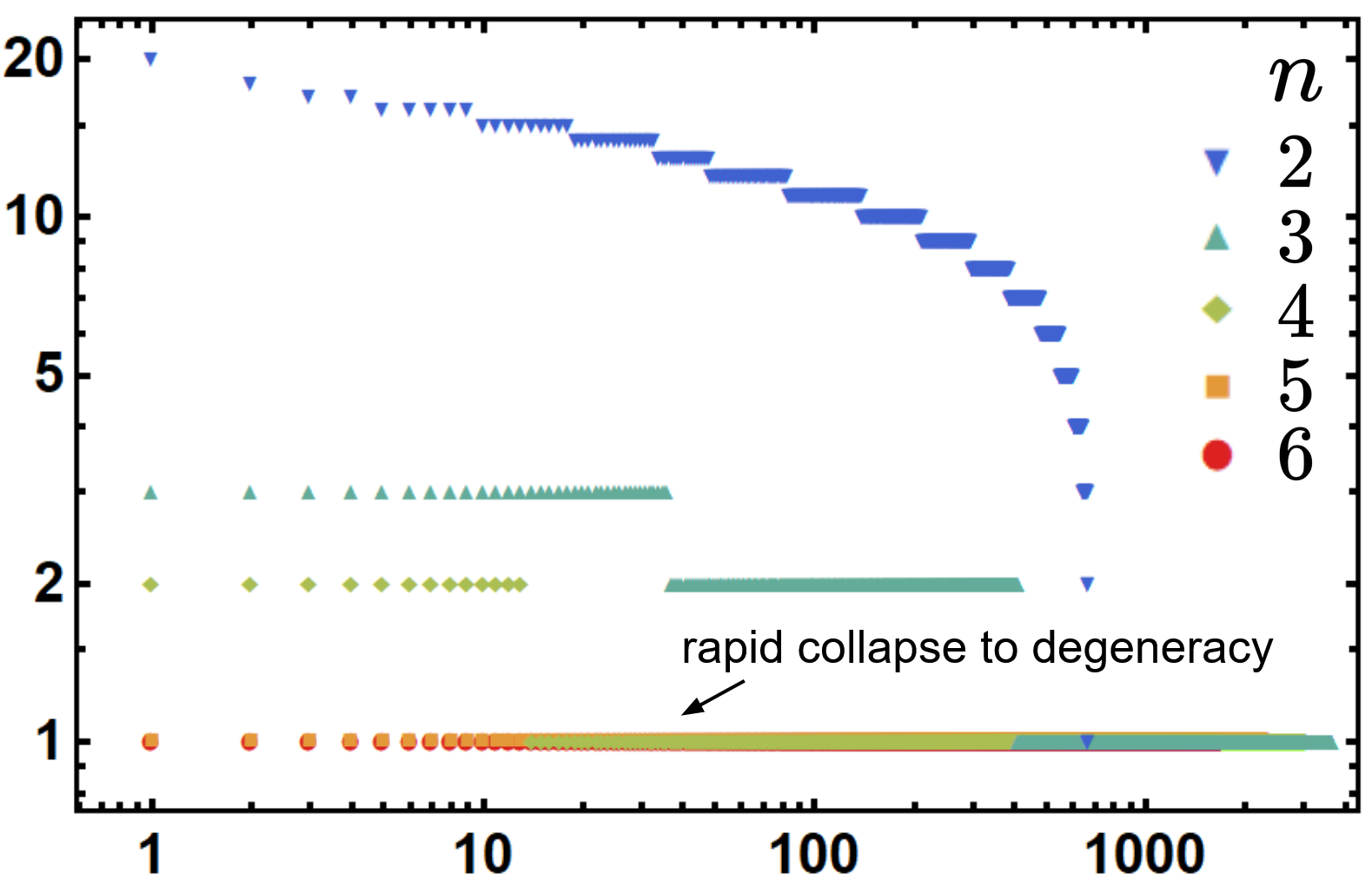}
    \caption{10000 character random string pulled from uniform distribution}
    \label{Fig:RandomStrings}
\end{subfigure}
\caption{Rank-ordered frequency distribution of $n$-grams}
\end{figure}

The random languages generated by our model exhibit this behavior because it generates smooth words -- meaning that words deeper in the hierarchy are composed only of $n$-grams at levels below it. This imbues the random language with a type of pseudo-morphology. By contrast, the frequency distribution of purely uniform strings (with or without spaces) quickly turns into a large degeneracy beyond $n_{\text{max}}$ (see Fig \ref{Fig:RandomStrings}). This is because the probability of generating a uniformly-random string of length $n$ falls exponentially, as $1/d^n$.

We briefly note that before $n_{\text{peak}}$, the frequency curves for both uniform strings (Fig \ref{Fig:RandomStrings}) and language data (Fig \ref{Fig:Alice}-\ref{Fig:yjjfgsp}) exhibit similar behaviors. In this initial regime, the frequency curve is predominantly power-law with an exponential tail. The power-law arises due to the fact that the small dimensions at early $n$ are saturated to their max value by the large number of random statistics. We find that this exponential tail vanishes at some $n$ just beyond $n_{\text{peak}}$. For {\it Alice} this transitions occurs between $n=3-4$, and $2-3$ for {\it yjjfgsp}.

\begin{figure}
    \centering
    \includegraphics[width=1.\linewidth]{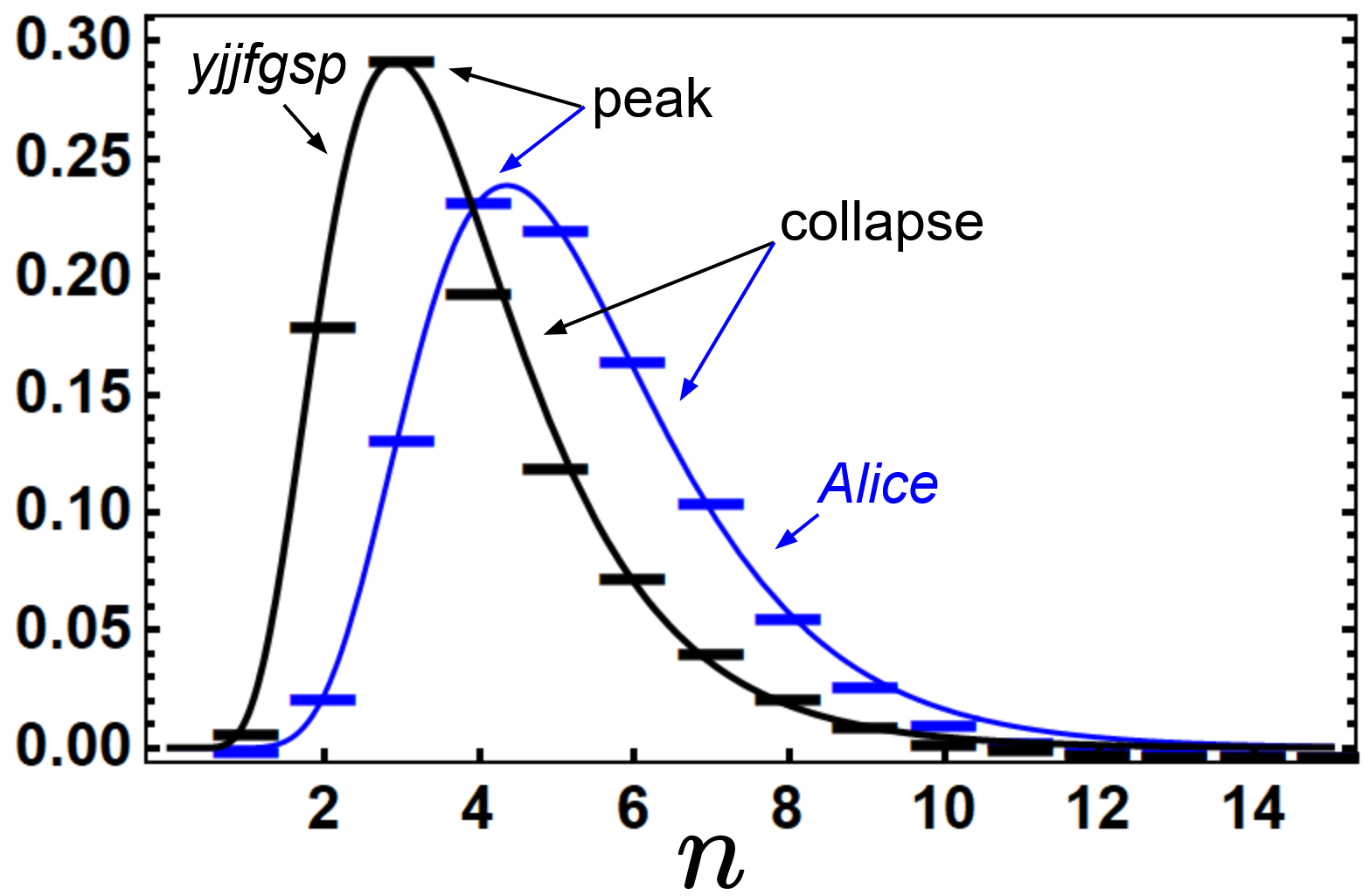}
    \caption{Comparing the normalized distribution of $n$-grams between ${\it Alice\, in\, Wonderland}$ (full text) and the random language {\it yjjfgsp}. The unnormalized distribution of {\it yjjfgsp} is $F(n,1.25,.43,2830)$.}
    \label{Fig:peak&collapse}
\end{figure}

Our random language was created {\it without} any training data. We did this by pulling the elements of $g_{\mu_{n-1},k}^{(n)}\in[0,1]$ from a uniform distribution. This is done successively for increasing $n$, where the $n$-grams with $g_{\mu_{n-1},k}^{(n)}>\epsilon_n$ are used to define the tokenset at $n+1$, same as discussed in Sec \ref{Sec:model}. Note that fine tuning of $\epsilon_n$ is not necessary in order to generate realistic distributions. We find that taking $\epsilon_n>0$ for the initial layers of the hierarchy is sufficient to guarantee collapse. For Fig \ref{Fig:yjjfgsp} \& \ref{Fig:peak&collapse}, we used $\epsilon_2=.7$, $\epsilon_3=.85$, $\epsilon_4=.45$, \& $\epsilon_{n>4}=0$.

Without any constraints, the effective dimension would scale exponentially as $d_n = d^n = e^{n\log(d)}$. This is because each of the $d$ initial basis tokens generates $d$ many compound tokens at the next scale. Those then $d^2$ tokens generate $d$ more tokens each, giving $d^3$, and so forth. Each token is the base of a branch which grows exponentially. By cutting one of these branches, we remove its contribution to the overall scaling exponent for the total dimension. This is the case initially at $n=2$ (for $\epsilon_2>0$), however additionally, the tokens at $n>2$ are further constrained by the requirement that they be smooth. We find that choosing sufficient $\epsilon_3>0$ is sufficient to cause the effective dimension to collapse at some later $n$, even if $\epsilon_{n>3}=0$. This is because some $n$-grams are necessarily terminating, in that it is impossible to add characters to it while remaining smooth. Such a collapse is demonstrated for the {\bf aba} toy language in Fig \ref{Fig:AliceHierarchy}-b. The dimensions of the hierarchy scale in two ways without fine tuning: explode or collapse. The fact that there is a largest word arises due to the collapse of this hierarchy.

While the phonotactics of some languages allow for longer word lengths (e.g some exotic constructions in Turkish), generally word length is not indefinite. Here we see that model parameters ($\epsilon_n$ \& $\tau_g$) place a limitation on word length. (In Appendix C, we discuss how $\tau_g$ \& $\epsilon_n$ are related.) If such limitations did not exists, then we would observe at least one language which has evolved extremely long words for typical conversation \cite{Herdan1958_logNormal_intraword,MarkNewman2006_powerlaws}. Rather, without exception, human language is universally hierarchical. More precisely, it is a hierarchy of hierarchies, which we assume arises out of the practical need to communicate longer strings of information than is allowed by the collapse of the first (intra-word) hierarchy.

A different route to generating a random language is to train against a uniform string (no spaces). We found that similar principles to that described above govern their distributions, but with the added constraint that the chance of observing a given $n$-gram imprinting into memory falls off exponentially. Thus the randomly grown hierarchies (by sampling $g_{\mu_{n-1},k}^{(n)}$) have the benefit of exploring the allowed shape of memory, as determined by the model parameters, and unburdened by insufficient statistics. We go into more detail for these methods in the supplement.

We end by pointing out that realistic random language data can be generated from stochastic processes \cite{Zanette&Montemurro2002}. These methods do not simulate language learning or offer microscopic descriptions but instead provide a useful effective description based on language structure. Notably, it shows the dependence of language structure on scale, sample size, and context \cite{Zanette&Montemurro2002,Montemurro&Zanette2002_sampleSize}. Our model's distributions most closely match small samples (a few hundred unique words) of natural language (Fig \ref{Fig:Alice500words}-\ref{Fig:yjjfgsp}), suggesting that speech patterns arise from phenomena at multiple scales, with the smallest scale influenced by brain hierarchy limitations.

\begin{figure}
    \centering
    \includegraphics[width=1.025\linewidth]{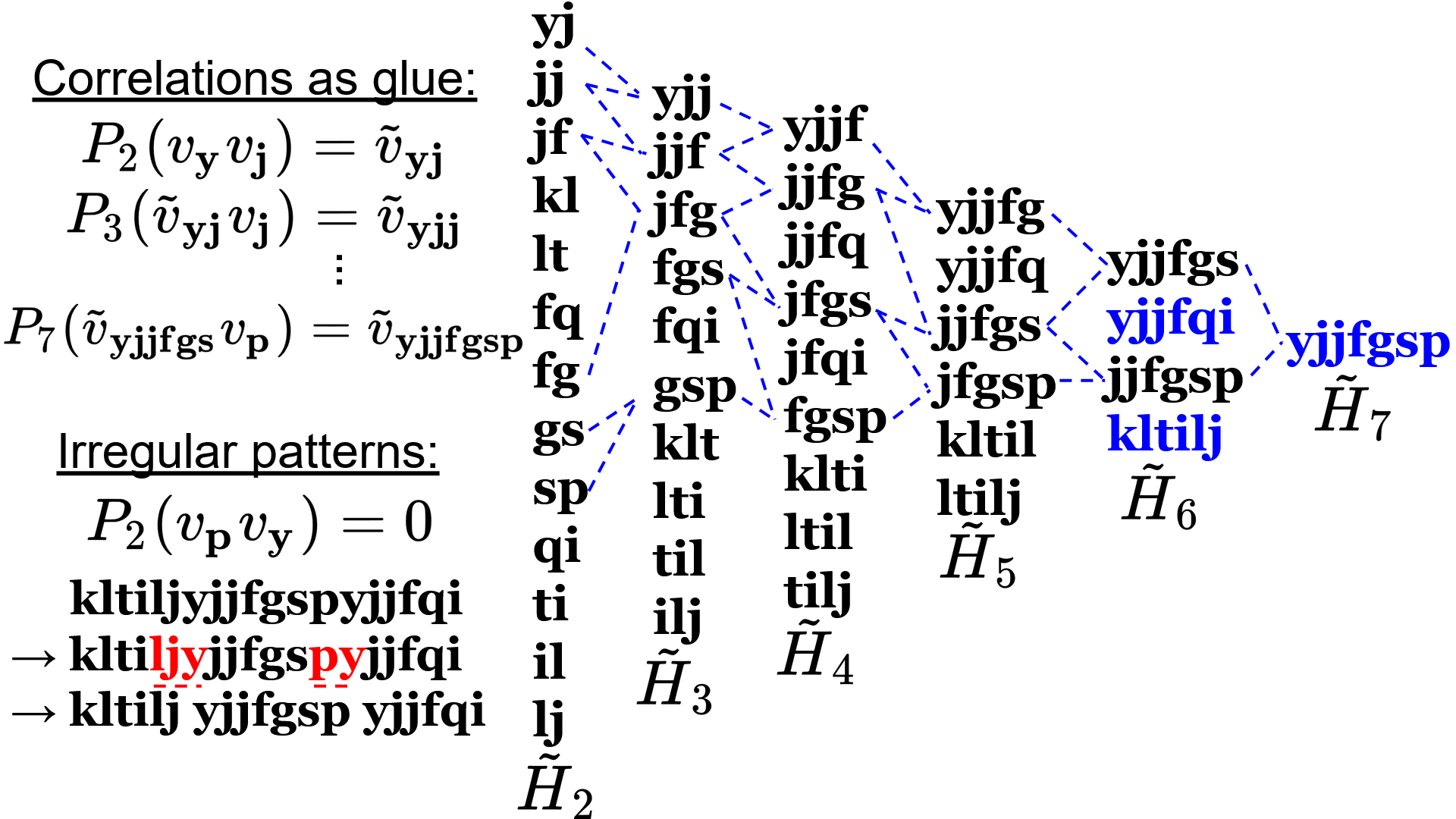}
    \caption{Summarizing the relationship between tokenizability, locality, and hierarchy.}
    \label{Fig:RandHierarchy}
\end{figure}

\section{neural morphology}\label{Sec:neuralmorphology}

Smooth tokensets shaping language structure arise from universal neural constraints, revealing that morphology is both a neural code and an inherent property of neural coding. This could explain why sign languages develop morphology \cite{Aronoff_SignLanguageMorphology}. Retinal data fits to maximum entropy models already suggest hierarchical, modular connections \cite{Schneidman2011_Sparse,Schneidman2020_RandomProjections}. Identifying a smooth tokenset could reveal fundamental neural building blocks.

One approach is fitting a maximum entropy model, then performing a greedy inference, which infers smooth tokens by iteratively selecting neurons that minimize the effective energy. This technique exploits locality to find smooth tokens with linear complexity.

Crucially, these models must be fit in a way which does not ruin the locality. Random projection models \cite{Schneidman2020_RandomProjections}, structurally similar to the projectors used here (see next section), offer a biologically plausible training method and have have succeeded in capturing the correlations of hundreds of neurons from undersampled data.

A potential difficulty in this analysis is discerning intra-layer ($v_jv_k$) from inter-layer ($v_j\tilde{v}_{\mu_2}$) correlations. Maximum entropy models fail to capture temporal correlations \cite{Beggs2010review_MaximumEntropy&Time}, so non-dynamical methods may be necessary in distinguishing features from their correlations. Perhaps the scaling relationships (e.g Fig \ref{Fig:yjjfgsp}) offer a signature for establishing an interrelationship. A full analysis is left for future work.

\section{inter-layer assumptions}\label{Sec:randomprojections}

In this section we scrutinize our assumption that $P_{\mu_n,\mu_{n-1},k}^{(n)}$ does not violate the microscopic locality. Unlike the $g_{\mu_{n-1},k}^{(n)}$, which are trained in an explicitly dynamic and local fashion (following Eqn \ref{Eqn:g(t+dt)}), the projectors are learned by proxy; i.e where the most significant correlations $g_{\mu_{n-1},k}^{(n)}>\epsilon_n$ define the learned tokens at the next scale. In this way, we assume the projectors based on the relevant patterns of the previous layer, by treating an effective ``sensitivity" $\epsilon_n$ to those patterns over the learning time. This process is unsupervised, but leads to $3$-point correlations, since both $\tilde{v}_{\mu_{n-1}}$ AND $v_k$ need to be firing in order for $\tilde{v}_{\mu_n}$ to fire. 

Such a $3$-point interaction can be accounted for by a feed-forward connection with a firing threshold. For example, $\tilde{v}_{\bf run}^{(3)}$ has activation function $\tilde{v}^{(3)}_{\bf run}(x)=\theta(\tilde{v}_{\bf ru}^{(2)}(x)+\tilde{v}_{\bf un}^{(2)}(x+1)-\phi)$; then it is possible for the threshold $\phi$ to be larger than either neuron can individually impose. This effectively turns the sum (an OR-operation) into an AND, justifying the projector. Translational invariance, plus the fact that {\bf ru} \& {\bf un} overlap guarantees smoothness. This generalizes to later layers as $\tilde{v}_{\mu_n}^{(n)}(x) = \theta\big(\tilde{v}_{\mu_{n-1}}^{(n-1)}(x)+\tilde{v}_{\mu_{n-1}'}^{(n-1)}(x+1)-\phi\big)$, which are shown as the blue feed-forward connections in Fig \ref{Fig:hierarchicalchain}.

How these feed-forwards are dynamically learned is a point we leave to future research. The form of the proposed solution suggests a more generic form,  
\begin{eqnarray}
&&v_{\nu}^{(n)}(x)=\\
&&\sum_{\mu_{n-1}\mu_{n-1}'k}\lambda_{\nu,\mu_{n-1},\mu_{n-1}'}^{(n)}\,\theta\big(\tilde{v}_{\mu_{n-1}}^{(n-1)}(x)+\tilde{v}_{\mu_{n-1}'}^{(n-1)}(x+1)-\phi\big) , \notag
\end{eqnarray}
where $\lambda_{\nu,\mu_{n-1},k}$ is learned, and assume random $\phi\in(\Lambda_v,2\Lambda_v]$. This suggests features could be learned as a random projection \cite{Schneidman2020_RandomProjections} of neurons in the previous layer. Here the number of projected neurons being limited to $2$. Note the general activation function of a biological neuron can be more complicated \cite{Sardi&Vardi&all2017nature_BioNeuronsMultipleThresholdUnits}; but this minimal example demonstrates how learning could technically occur via a single set of parameters ($\lambda_{\nu,\mu_{n-1},\mu_{n-1}'}$), reducing the assumed computational load expected of a single neuron.

%We are uncertain what the true timescales are for learning these projections, nor their decay time, but it's not necessarily identical to $\tau_g$.

\section{discussion}
%Random hierarchy... \cite{Clauset&Moore&Newman2008nature_HierarchicalNetworks}

Physical reality is largely governed by principles which are local \& unsupervised. If organization happens, it happens absent any notion of correctness, arising instead on accident. This is opposite the semi-supervised paradigm for pre-training LLMs, where a global notion of correctness is prescribed in a loss function, which is minimized in order to bring the network behaviour inline with the desire one. The loss function (cross-entropy, KL-divergence, etc) is a conscious choice made by the modeler.

But the microscopic environment that governs decision making \& learning in humans is entirely unconscious. While a human learner can define any behaviour to be ``correct", how that arises from the 2-neuron level is not yet understood, let alone how the local components of the brain coordinate to achieve it. A neuron is not psychically aware of all other neurons elsewhere in the brain. 

Models of biological learning which assume correctness ignore this issue; and consequently take the form of black boxes which successfully fit the data, without explaining its structure or accounting for its origin. Children do not learn language by fitting the loss, but by building an unsupervised {\it interpretation} of the data. The unsupervised learning of the language is possible because of its local structure \cite{Solan&Horn&Ruppin&Edelman2005_ADIOS}, which we explain here as being due to a sparse \& local microscopic structure. Our model provides this explanation without needing to fit already existing data, which is a scientific necessity in order to account for why language data exists in the first place.
%Our model not only fits the data but explains it. 

The approach taken here follows a constraints-driven minimal approach used in theoretical physics. The constraints place limitations on how the observables at different scales can be connected. We then asked what assumptions (i.e structures) are necessary to make this connection, e.g the hierarchies. These hierarchies provide an explanation, but suffer from a clear computational limitation. This explanation is therefore either incorrect or incomplete. Assuming the latter means that this limitation is one that the brain overcomes on accident. This led us to assume the random pinning field $\psi$ (in Sec \ref{Sec:replay}), which we stress is not meant to be taken literally, but is a convenience for exploring such accidents. If while a hierarchy is performing inference, we hold a neuron close to the hierarchy (i.e pin it), it does a useful computation. We then throw that neuron onto a pile of trained embedding neurons, and repeat the game with another neuron taken from a source at infinity.

This pinning of the neuron guarantees an effective {\it simultaneity} for the replayed $n$-grams. In the brain, this notion of simultaneity possibly arises as a spatio-temporal consequence of the structure and connectivity of the brain; and may additionally depend on the statistics of avalanches \cite{Beggs&Plenz2003_NeuronalAvalanches,Friedman&Ito&all&Beggs2012prl_Criticality,Fosque&WilliamsGarc&Rashid&Beggs&Ortiz2021prl_Quasicriticality,Gregory&all2014_MouseCortexCriticality}, which can carry simultaneous firing over longer distances \cite{Gregory&all2014_MouseCortexCriticality,Cocchi&Gollo&Zalesky&Breakspear2017_reviewCriticalityBrain,Plenz&all2021_SelfOrganizedCriticality}.

%The structures of the brain arise due to the environmental conditions of development \cite{Stiles&Jernigan2010review_BrainDevelopment}.

Note that simultaneity can be used to tie together visual and spoken features, e.g: {\it Want to learn the word {\bf girl}? Then say {\bf girl} and imagine a girl. If there some neuron firing during the replay of all those feature sets, then it becomes correlated with those feature sets.} We would then write 
$H_{\text{\bf girl}} = a_{{\bf girl}}(\sum_{\chi}b_{{\bf girl},\chi} + \sum_n\sum_{\mu_n}m_{{\bf girl},\mu_n}^{(n)}\tilde{v}_{\mu_n})$, where $\sum_{\chi}b_{{\bf girl},\chi}$ is the placeholder for the sum over the visual features. The token $a_{\bf girl}$ now carries ``meaning" as defined by its feature set. This example demonstrates how symantics can arise through muscle memory, i.e Hebbian processes.

A likely candidate for the location of the embedding neurons is Broca's area (BA 44 \& 45), as well as Wernicke (BA 22). In a recent meta-analysis comparing activated brain regions of sign-language and spoken-language speakers, Broca's area was the notable overlap \cite{Trettenbrein&Papitto&Friederici&Zaccarella2021_MetaAnalysisBrocas}. Likewise, Broca and Wernicke areas are expected to be tied to the symanticful content of speech \cite{Kemmerer2015book_CognitiveNeuroscienceOfLanguage}. Patients suffering from lesions in Wernicke's area suffer {\it fluent aphasia}, where structural fluent but meaningless speech is produced \cite{Binder2015review_WernickeArea,Kemmerer2015book_CognitiveNeuroscienceOfLanguage}. Lesions in Broca's area can lead to complete loss of speech \cite{Stinnett&Reddy&Zabel2023_BrocasNeuroanatomy,Kemmerer2015book_CognitiveNeuroscienceOfLanguage}.

This distinction between symantics-free ($\tilde{v}$) \& symantics-full ($a$) networks explains the two effects experienced by the reader when parsing the strings of Fig \ref{Fig:savevile}. The first being the \underline{tokenizability}, where word boundaries can be located by looking for non-word forming correlations. This should be understood as being a {\it property of the language}, which evolved within the limitations of the tokenizing hierarchies. The second effect is \underline{recognition}, where part of a familiar word triggers our learned embedding for that word. We exploited this fact in order to trick the reader attempting to tokenize the final two strings of Fig \ref{Fig:savevile}. Fast embedding neurons pick up familiar words living across the word boundaries, which is a distraction from finding the two words which uniquely tokenize the string.

We argue that replay relearning describes imaginative human thought. Such {\it thinking} does not need to be taught to the model, rather it arises as an accident due to the conditions at the microscopic level. Neurons learn off the random replay of other neurons, and from this dynamical process emerges a {\it key-value} memory 
\cite{Gersham&Fiete&Irie2025_KeyValueMemory}, Eqn \ref{Eqn:vvmv}, which can be understood as a simple attention mechanism. These disentangled embeddings can then be used to form their own hierarchies, $\sum_{m}\tilde{a}_{\alpha_m}a_{\alpha}$, except now these hierarchies describe the interactions between symanticful tokens. 

We infer the existence of this term because it is allowed by locality. For a given set of starting assumptions (e.g the number of hierarchies), a finite number of locality-preserving terms exist. The biological plausibility of these terms justifies their systematic study, which includes exploring the order and manner that these networks learn off each other. As discussed in Sec \ref{Sec:ralang}, the shape of memory is a leading order effect on the structure of language. Thus the distributions which govern the patterns of speech may act as a source of insight \& data for constraining an effective human language model. For example, fitting to transcribed samples from patients with aphasia. The mathematical framework worked out in this manuscript makes possible this exploration.

\section{conclusion}

In this paper, we provide an answer to the question: What is the microscopic origin of the local correlations in language? Without exception, human language is universally {\it local} and {\it hierarchical}, and continues to remain so despite generational drift. We argue that locality arises due to the local nature of the microscopic neuron-neuron coupling. This places a strong limitation on the brains ability to produce correlated strings of even moderate length, which it does by forming a predictive hierarchy. These hierarchies learn unsupervised a tangled series of projection maps, which are needed to tokenize the data. Other neurons can then learn off the replay of these hierarchies, by tying the replayed features to an embedding. This disentangles the projection map, making possible both a significant compression and continual parallel learning. We argue that the tokenizable patterns which constitute morphology are a reflection of a tokenizable neural code, which has a distinct scaling signature (see Fig \ref{Fig:yjjfgsp}, \ref{Fig:peak&collapse}, \& \ref{Fig:RandHierarchy}) that we predict can be found in neural data.

\subsection*{Appendix A}\label{Appendix:A}

We'll now establish the stability of Eqn's \ref{Eqn:Hebb} \& \ref{Eqn:g(t+dt)} during training. It will be sufficient to show scalar function $g(t)$ does not explode under evolution by $\tau_g\dot{g}+g=\Lambda$. Here $\Lambda$ is a constant representing the effect of pinning both neurons high during training. A general solution to the ODE follows $g(t) = g(0)e^{-t/\tau_g}+\Lambda$, which is bounded. Information is quickly forgotten for times $t\ge\tau_g$.

\subsection*{Appendix B}\label{Appendix:B}

Here we detail how to perform $\mathcal{L}_n$. For simplicity, it will suffice to drop the $x$ index and write $P_n = \sum\limits_{\mu_n,\mu_{n-1},k}P_n^{\mu_n,\mu_{n-1},k}\tilde{v}_{\mu_n}^{(n)}\tilde{v}_{\mu_{n-1}}^{(n-1)}v_k$, with the understanding that index order determines position. (Note for tensors like $\sum_{jk}T_{jk}v_j(x-1)v_k(x)\equiv \sum_{jk}T_{jk}v_jv_k$, it should be understood that $v_jv_k$ do {\it not} commute, and $T_{jk}\neq T_{kj}$ in general.) We will need to construct a regauged set of projectors of the form 
\begin{eqnarray}
P_n^r = \sum\limits_{\mu_n,\mu_{n-1},k}P_n^{r;\mu_n,k,\mu_{n-1}}\tilde{v}_{\mu_n}^{(n)}v_k\tilde{v}_{\mu_{n-1}}^{(n-1)} . 
\end{eqnarray}
$P_n$ \& $P_n^r$ are different representations of the same object. One way to construct it is to left-tokenize $g^{(3)}_{\mu_2,k}\rightarrow g^{(3)}_{k,\mu_2}\equiv\sum_{lz}P_2^{\mu_2,l,z}\sum_{\mu_2'}P_2^{\mu_2',k,l}g^{(3)}_{\mu_2',z}$. (Note in our notation $g^{(3)}_{k,\mu_2}\neq g^{(3)T}_{\mu_2,k}$. Indices should only be reordered by projector maps.) Then define $P_3^{r;\mu_3,k,\mu_2}=1$ using the $v_k\tilde{v}_{\mu_{2}}^{(2)}$ for which $g^{(3)}_{k,\mu_2}>\epsilon_3$ (and $0$ otherwise). We then use $P_3^r$ to left-tokenize $H_4$: $g^{(4)}_{\mu_2,l,k}\equiv \sum_{\mu_3}P_3^{\mu_3,\mu_2,l}g^{(4)}_{\mu_3,k}$, then $g^{(4)}_{\mu_2,\mu_2'}=\sum_{lk}g^{(4)}_{\mu_2,l,k}P_2^{\mu_2',l,k}$, then $g^{(4)}_{j,k,\mu_2}\equiv \sum_{\mu_2'}g^{(4)}_{\mu_2',\mu_2}P_2^{\mu'_2,j,k}$, and finally $g^{(4)}_{k,\mu_3}\equiv\sum_{\mu_2,j}g^{(4)}_{k,j,\mu_2}P_3^{r;\mu_3,j,\mu_2}$. Then define $P_4^r$ using the $v_k\tilde{v}_{\mu_{3}}^{(3)}$ for which $g^{(4)}_{k,\mu_3}>\epsilon_4$. Repeat this process until you left-tokenized $H_n$, which requires $P_{m}^{r}$ for all $m<n$.

\subsection*{Appendix C}\label{Appendix:C}

Here we demonstrate how to decode the learned compound tokens into the basis set. Consider the example token $\tilde{v}_{\bf would}$, which is a $5$-gram. Apply the inverse of the projector $P_5^{T}(\tilde{v}_{\bf would})=\tilde{v}_{\bf woul}v_{\bf d}$. The final token $v_{\bf d}$ can be peeled off using an SVD decomposition, $\text{SVD}(\tilde{v}_{\bf woul}v_{\bf d})=\{U,D,V\}$, where $UD=\tilde{v}_{\bf woul}$ and $DV=v_{\bf d}$. Thus we can rewrite $\text{SVD}(\tilde{v}_{\bf woul}v_{\bf d})=\{\tilde{v}_{\bf woul},v_{\bf d}\}$. Then start again with $\tilde{v}_{\bf woul}$. Repeat this to produce the list $\{v_{\bf w},v_{\bf o},v_{\bf u},v_{\bf l},v_{\bf d}\}$.

\subsection*{Appendix D}\label{Appendix:D}
Here we derive a relationship between the cutoff $\epsilon_n$ and $\tau_g$. Starting from Eqn \ref{Eqn:g(t+dt)}, we see that a single $t\rightarrow t+dt$ instance of $\tilde{v}_{\mu_{n-1}}^{(n-1)}v_k>0$ imprints an $\mathcal{O}(\xi_g)$ contribution to $g_{\mu_{n-1},k}^{(n)}$. In order for $g_{\mu_{n-1},k}^{(n)}>\epsilon_n$, at least $M>\epsilon_n/\xi_g$ instances of $\tilde{v}_{\mu_{n-1}}^{(n-1)}v_k>0$ need to observed within $\tau_g$. Since $\xi_g=N_g^{-1}$, we can equivalently write $M/N_g>\epsilon_n$. Thus $\epsilon_n$ defines a lower bound on the frequency, below which the model is insensitive.

\subsection*{Appendix E}\label{Appendix:E}
Fits of \ref{Fig:AliceHierarchy} \& \ref{Fig:peak&collapse} follow the log-normal distribution \cite{Herdan1958_logNormal_intraword,Williams1940_logNormal_interwords,Eckhard&Werner&Markus2001_logNormal_review}, 
\begin{eqnarray}
F(n,\mu,\sigma,\mathcal{N})=\mathcal{N}\frac{\exp\Big(\frac{-(\log n - \mu)^2}{2\sigma^2}\Big)}{n\sigma\sqrt{2\pi}} . 
\end{eqnarray}

%\subsection{Appendix D}
%The formalism adopted in the main text employs a semi-dense structure composed of projector maps $P_n$ and Hamiltonians $\tilde{H}_n$ which are dense matrices. An alternative fully-sparse fit of Eqn \ref{Eqn:IsingN} is discussed here.

\subsubsection{acknowledgments}

Much thanks to John Beggs, Dimitry Krotov, Paul Boersma, Zach Solan, Ricky Simanjuntak, Ceren Da\u{g}, and Luke Denny for discussion. 

The compression code was written using Julia ITensor \cite{ITensor}. The solutions to Fig \ref{Fig:savevile}: know+yes, orbital+sandwich, train+hello, save+vile or ave+viles, monk+eye, runt+see, sewing+oat, \& sold+read or old+reads or dread+sol.

Yjjfgsp: \cite{yjjfgsp}

\bibliography{biblio} 
\end{document}